\documentclass{article} 
\usepackage{iclr2027_conference,times}

\usepackage{amsmath,amsfonts,bm}

\def\eqref#1{equation~\ref{#1}}

\def\1{\bm{1}}

\newcommand{\test}{\mathcal{D_{\mathrm{test}}}}

\DeclareMathAlphabet{\mathsfit}{\encodingdefault}{\sfdefault}{m}{sl}
\SetMathAlphabet{\mathsfit}{bold}{\encodingdefault}{\sfdefault}{bx}{n}

\usepackage{hyperref}
\usepackage{amsmath,amssymb,booktabs,graphicx,makecell,multirow,longtable,array}
\usepackage{microtype}
\usepackage{subcaption}
\usepackage[ruled,vlined,algo2e]{algorithm2e}
\usepackage{url}
\usepackage{placeins}
\title{Are We Recovering Mechanisms?\\
Objective-Level Recovery Gaps in Mechanistic Interpretability}

\author{Chuqin Geng \\
University of Toronto \\
McGill University \\
\texttt{chuqin.geng@mail.mcgill.ca}
\And
Li Zhang \\
University of Toronto
\And
Haolin Ye \\
McGill University
\AND
Mark Zhang \\
University of Toronto
\And
Luke Zhang \\
University of Toronto
\And
Xujie Si \\
University of Toronto \\
\texttt{six@cs.toronto.edu}
}

\iclrfinalcopy 

\newsavebox{\appendixtablebox}

\begin{document}
\maketitle
\lhead{Preprint}
\raggedbottom

\begin{abstract}
Mechanistic interpretability aims to recover the internal computations responsible for model behavior. Progress in automated circuit discovery is often framed as a search problem: better attribution or optimization should identify better mechanisms. This assumes that the evaluation objective can recognize a better circuit once it is found. Yet standard intervention-based evaluation replaces signals outside a circuit to isolate its effect, changing the computational environment in which the retained components operate. We call this change \emph{context distortion}. This raises the question of whether faithfulness rewards recovery of model behavior or compatibility with the altered context. We show that intervention-based faithfulness can prefer an equally sized circuit that reproduces the model's behavior less well. We call this reversal \emph{misranking}, an \emph{objective-level recovery gap}. To isolate evaluation from discovery, we construct equally sized candidates through controlled edits to human-identified circuits in pretrained language models, including IOI and Greater-Than, and benchmark circuits with known mechanisms. We find misranking even without a discovery algorithm. Misranking also occurs in real automated circuit-discovery settings. Among outputs of EAP, EAP-IG, ACDC, and Edge-SP, KL faithfulness under ordinary resampling prefers the behaviorally worse candidate in 13.2\%--24.4\% of candidate pairs across these methods on four pretrained-language-model tasks. Finally, restoring selected intact-model signals reverses 96 of 100 selected persistent misrankings, demonstrating that context distortion contributes to these failures. In light of these findings, we call for rethinking both how circuit discovery is performed and how its results are evaluated.
\end{abstract}

\section{Introduction}
\label{sec:introduction}
Mechanistic interpretability seeks to identify the computations behind model behavior. Circuit analysis represents a computation as a sparse subgraph of interacting model components \citep{olah,framework}. Detailed studies have reconstructed circuits for behaviors such as indirect object identification (IOI) and numerical comparison \citep{wang,gt}, while automated methods aim to find such circuits at scale \citep{conmy,syed,hanna}. This work often treats recovery as a search problem: better attribution, more data, or improved optimization should find a better account of the computation used by the intact model. But better search only helps if the evaluation objective can recognize and favor the better circuits it finds.

In practice, circuits are typically evaluated through intervention-based \emph{faithfulness}: how well they preserve model outputs when computation outside them is replaced \citep{conmy,hanna}. Faithfulness guides both search and evaluation. It can determine which connections are retained and which candidate is ultimately presented as an explanation. A high score is therefore a reason to choose one circuit over another. This raises the question we study:
\begin{center}
\emph{Can faithfulness prefer an equally sized circuit that reproduces the model's behavior less well?}
\end{center}

\begin{figure*}[!t]
  \centering
  \includegraphics[width=\linewidth]{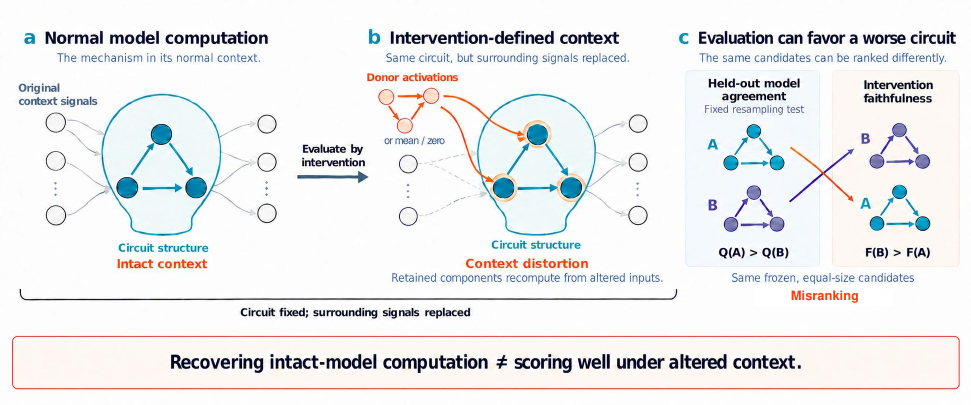}
  \caption{\textbf{The objective-level recovery gap.} (a) Retained components (blue) compute in the intact model\textquotesingle s context. (b) The same circuit receives donor, mean, or zero replacements (orange) on excluded connections and recomputes under altered inputs. (c) A and B are two distinct, equally sized candidates. Held-out behavioral performance $Q$ under fixed ordinary resampling prefers A, but intervention-defined faithfulness $F$ prefers B; higher values are better for both. The crossing arrows show this reversal. Panels (a)--(b) change one circuit\textquotesingle s context; panel (c) compares two circuits.}
  \label{fig:intro}
\end{figure*}

Intervention-based evaluation replaces signals outside a circuit with donor activations, means, or zeros to isolate the circuit's effect \citep{miller,patchguide}. Yet this replacement also changes the computational environment in which the circuit operates. In particular, retained components can receive different inputs and recompute their activations under conditions unlike those of the intact model. We call this change \emph{context distortion}. If two circuits respond differently to that altered environment, faithfulness can favor the one that performs better under intervention even when it reproduces intact-model behavior less well. Figure~\ref{fig:intro} illustrates this wrong preference.

This is an \emph{objective-level recovery gap}: a search method can return a better circuit, yet the score used to choose among its outputs can penalize it. We study the gap through pairwise \emph{misranking}, when validation faithfulness favors the circuit with worse held-out behavior. We measure held-out behavior under a fixed ordinary-resampling protocol shared across candidates.\footnote{Greater-Than uses semantic completion accuracy: a predicted two-digit year ending must exceed the starting ending. All other tasks use intact-model answer agreement, including the model's mistakes.} We compare circuits of the same size that are already available, asking whether the score picks the one that better reproduces model behavior. In the most striking cases, a behaviorally worse candidate even scores higher than a benchmark's known-mechanism reference circuit.

We investigate this problem across two complementary settings: semi-synthetic models with known mechanisms and pretrained language models with previously studied circuits. InterpBench provides known mechanisms in smaller, semi-synthetic models \citep{gupta}. Four tasks in pretrained language models have previously studied circuits, including IOI \citep{wang} and Docstring \citep{docstring}; we call these the Human suite. We use both suites to ask whether evaluation itself can misrank circuits, whether the failure occurs among outputs of discovery methods, and whether restoring computational context reverses the wrong choice. We find:
\begin{itemize}
\item \textbf{Evaluation can prefer a worse circuit without any discovery algorithm.} Controlled edits produce equally sized pairs for which faithfulness favors the circuit that reproduces intact-model behavior less well. Because no discovery algorithm generates these candidates, the reversal arises in evaluation itself (Section~\ref{sec:controlled}).
\item \textbf{The same failure can steer real circuit discovery.} Misranking occurs among outputs of EAP, EAP-IG, ACDC, and Edge-SP. Under ordinary resampling, KL faithfulness prefers the behaviorally worse candidate in 13.2\%--24.4\% of candidate--candidate pairs across these methods on the four pretrained-language-model tasks. Misranking is a recurring phenomenon across faithfulness metrics, occurring under KL, logit-difference, and probability-difference scores (Section~\ref{sec:discovery}).
\item \textbf{Context distortion contributes to the wrong preference.} Intervention-based evaluation replaces signals outside a circuit, thereby changing the computational environment in which it operates. Restoring selected intact-model signals reverses 96 of 100 selected persistent misrankings. The result points to the intervention context as a reason faithfulness can reward the wrong circuit, and to the need to rethink how circuits are discovered and evaluated (Section~\ref{sec:repair}).
\end{itemize}

\section{Preliminaries}
\label{sec:framework}
\subsection{Tasks, reference circuits, and interventions}
We study circuit rankings in two settings: pretrained language models with circuits identified in prior work, and semi-synthetic models with known computations (Table~\ref{tab:tasks}). We refer to the existing circuit in each task as its \emph{reference circuit}. For the \emph{Human suite}, we use the circuits from published analyses of IOI, Greater-Than, Docstring, and Acronym. For InterpBench, we use the computation supplied with each semi-synthetic model \citep{gupta}. These references give us a starting point for controlled edits and a size budget for discovery. We evaluate their behavior alongside the candidates rather than assuming that a reference must always perform better.

\paragraph{Circuit representation and task selection.}
A circuit specifies both the components involved in a computation and the connections through which they exchange information \citep{conmy,edgepruning}. Node masks alone cannot distinguish circuits that retain the same components but use different connections. Edge masks provide finer control over information flow by specifying which upstream signals each retained component receives \citep{edgepruning}. We therefore use edge masks to preserve the wiring differences among candidates produced by EAP, EAP-IG, ACDC, and Edge-SP. We use this representation for IOI, Greater-Than, and Docstring, while keeping Acronym's whole-head representation.

InterpBench requires an additional step because its native evaluator retains nodes. To express that computation in our edge evaluator, we keep every outgoing connection of each retained node. This matters: simply copying the listed graph edges can omit contributions that native node execution preserves. We check that each converted reference reproduces the native outputs under resampling, mean, and zero. We then select the ten largest categorical references that pass this check, ranked by their executable edge count. Appendix~\ref{app:section3} gives the selected tasks and the conversion checks.

\begin{table}[!b]
\centering
\caption{\textbf{Tasks and reference circuits.} Behavior names the prediction task; Reference size counts executable edges, except for Acronym, which counts attention heads; Source cites the published circuit analysis or benchmark. The Human suite comprises the first four tasks. InterpBench contains ten semi-synthetic tasks, whose reference sizes span the reported range. We match circuit size within every compared pair.}
\label{tab:tasks}
\small
\setlength{\tabcolsep}{5pt}
\begin{tabular*}{\linewidth}{@{\extracolsep{\fill}}llrl@{}}
\toprule
Task & Behavior & Reference size & Source\\
\midrule
IOI & Indirect-object name prediction & 963 edges & \citep{wang}\\
Greater-Than & Year-ending comparison & 235 edges & \citep{gt}\\
Docstring & Python docstring completion & 24 edges & \citep{docstring}\\
Acronym & Acronym letter prediction & 8 heads & \citep{acronym}\\
InterpBench (10) & Semi-synthetic computations & 72--587 edges & \citep{gupta}\\
\bottomrule
\end{tabular*}
\end{table}

\paragraph{Circuit interventions.}
Let $M$ denote the intact model, $S$ a circuit mask specifying the retained components, and $M_{S,a}$ the circuit executed under intervention $a$. We evaluate a circuit by replacing signals on its excluded connections as the model computes its forward pass. Under \emph{resampling}, we supply activations cached from another prompt, called the donor; under \emph{mean} ablation, we supply activations averaged over a prompt bank; under \emph{zero} ablation, we supply zeros. The retained components still compute normally, but they now receive some replaced inputs \citep{miller,patchguide}. Thus, keeping the circuit's structure fixed does not keep its computational context fixed.

\paragraph{Validation faithfulness.}
We first score each circuit on a validation set $V$. For a recipient prompt $x$, let $p_M(\cdot\mid x)$ and $p_{S,a}(\cdot\mid x)$ denote the intact model's and intervened circuit's output distributions. Our primary score $F_a(S)$ is their negative KL divergence:
\begin{equation}
F_a(S)=-\mathbb{E}_{x\in V}\!\left[D_{\mathrm{KL}}\!\left(p_M(\cdot\mid x)\,\|\,p_{S,a}(\cdot\mid x)\right)\right].
\label{eq:kl}
\end{equation}
A higher score means that the output distributions are closer. We also compute logit difference (LD) and probability difference (PD), which measure the margin between task-correct outputs and alternatives. We obtain all three scores from the same circuit executions, so changing the metric does not change the candidates or their intervention context. We give the score definitions and additional results in Appendices~\ref{app:protocol}, \ref{app:section3}, and~\ref{app:section4}.

\paragraph{Held-out behavior.}
A recovered circuit should reproduce the behavior of the model it explains. We therefore assess each circuit on a separate test set $T$, asking how well it reproduces intact-model answers on unseen prompts. This provides a behavioral criterion for comparing recovery quality, while allowing alternative circuits that reproduce the same behavior. We use \emph{fixed ordinary resampling} to measure this held-out performance, denoted by $Q$. Let $I_x$ be the scored positions of prompt $x$, and let $\hat y_M(x,i)$ and $\hat y_{S,\mathrm{resample}}(x,i)$ be the intact model\textquotesingle s and circuit\textquotesingle s answers at position $i$. For model-answer agreement, we compute
\begin{equation}
Q(S)=\mathbb{E}_{x\in T}\!\left[\frac{1}{|I_x|}\sum_{i\in I_x}
\mathbf{1}\!\left\{\hat y_{S,\mathrm{resample}}(x,i)=\hat y_M(x,i)\right\}\right],
\label{eq:q}
\end{equation}
The indicator $\mathbf{1}\{\cdot\}$ equals one when the two answers match and zero otherwise. We use the intact model\textquotesingle s answer as the target, including its mistakes. When we change the validation metric or intervention, we leave this test protocol unchanged. This gives us a fixed behavioral ordering against which to judge each faithfulness ranking. We apply the same test to the reference and every candidate.

Our results support this behavioral criterion: the ten InterpBench reference circuits achieve 97.33\%--100\% held-out agreement, and the published IOI circuit achieves 99.00\%. Moreover, mean candidate performance generally declines as we replace more reference components while preserving circuit size (Tables~\ref{tab:perf3Human4} and~\ref{tab:perf3Interp10}). Thus, $Q$ captures behavioral losses associated with disrupting these reference computations.

For each task, we use 100 validation prompts, 100 disjoint test prompts, and a separate 100-prompt bank for estimating mean activations. For resampling, we evaluate each recipient prompt with three donors and give every candidate the same donor assignments. We average over scored positions and donors within each prompt, then average over prompts. We reuse these evaluation sets and assignments in Sections~\ref{sec:controlled} and~\ref{sec:discovery}; the discovery methods receive separate prompts.

\subsection{Misranking as a failure of circuit selection}
For each pair, we call the circuit with higher held-out $Q$ the better circuit $B$, and the other the worse circuit $W$. We say that faithfulness \emph{misranks} the pair when
\begin{equation}
\underbrace{Q(B)>Q(W)}_{\text{held-out behavior prefers }B},
\qquad
\underbrace{F_a(W)>F_a(B)}_{\text{validation faithfulness prefers }W}.
\label{eq:misranking}
\end{equation}
We measure both how often this happens and how much behavior is lost by following the score. The gap $\Delta Q=Q(B)-Q(W)$ is the behavioral deficit of the preferred circuit, while $\Delta F=F_a(W)-F_a(B)$ measures how strongly faithfulness favors it. We report $Q$ as a percentage and $\Delta Q$ in percentage points (pp). Distribution similarity and answer agreement need not give identical rankings: a small change near a decision boundary can flip an answer, whereas a larger change elsewhere may leave the answer unchanged. Our question is whether this difference makes faithfulness choose behaviorally worse circuits in practice, and whether the resulting losses are substantial.

\emph{We match executable size in every primary comparison.} A smaller circuit might naturally preserve less behavior because it retains fewer components; size matching removes that explanation for the reversals we study. We make two kinds of comparison. Reference--candidate (R--C) pairs test whether faithfulness can choose correctly between a reference and an alternative. Candidate--candidate (C--C) pairs test whether it can choose between alternatives. In either case, we determine the better circuit from $Q$, not from its status as a reference. For discovery outputs, we also keep the task and method fixed and remove duplicate masks before forming pairs.

We compute misranking rates over all distinct eligible pairs, keeping score ties and $Q$ ties in the denominator. We average behavioral deficits over the strictly misranked pairs. To estimate uncertainty, we resample prompts while keeping each pair's measurements together (a paired bootstrap), using a common numerical tolerance (Appendix~\ref{app:protocol}). Because a circuit can appear in several pairs, these rates describe the candidate pools we evaluate.

\paragraph{Objective-level recovery gap.}
The goal of circuit discovery is to recover the internal computation responsible for intact-model behavior. In practice, discovery optimizes an intervention-defined faithfulness objective. Over a family $\mathcal{S}_K$ of size-$K$ circuits, this objective favors
\begin{equation}
S_{F,a}^{\star}\in\arg\max_{S\in\mathcal{S}_K}F_a(S).
\label{eq:objective-targets}
\end{equation}
But this objective evaluates circuits in a computational environment altered by intervention. Replacing excluded signals changes the inputs on which retained components compute, creating \emph{context distortion}. Faithfulness therefore rewards matching intact-model outputs within this altered environment. A circuit can score better because its computation is more compatible with the replacement signals, rather than because it better recovers the mechanism used during normal inference. We call this mismatch an \emph{objective-level recovery gap}.

Our held-out criterion $Q$ tests one requirement of recovery: reproducing the model's behavior on unseen prompts under a fixed resampling protocol. When faithfulness favors a circuit with worse $Q$ over an equally sized alternative, it fails this behavioral test of recovery. Improving search alone cannot correct that preference.

\section{Faithfulness Evaluation Alone Can Misrank Circuits}
\label{sec:controlled}
We first ask whether faithfulness evaluation alone can prefer a worse circuit. To isolate evaluation from discovery, we construct equally sized candidates through controlled edits to reference circuits and compare their faithfulness rankings with their held-out behavior. A reversal in this setting exposes a failure of evaluation itself: the better circuit is already available, yet the score favors the worse one.

\paragraph{Controlled candidate pools.}
We generate each candidate by removing $k$ reference components and adding $k$ excluded components. We consider one-component edits and replacements of 5\%, 10\%, 20\%, 50\%, and 75\% of the reference, using 30 seeds for each distinct attainable edit count. This gives us both small changes near the reference and more substantial alternatives, all at the same size. We generate the candidates before scoring, retain them regardless of performance, and merge duplicate masks and edit counts (Appendix~\ref{app:section3}).

\begin{table}[t]
\centering
\caption{\textbf{Faithfulness misranks frozen, equally sized circuits.} Columns group the validation intervention (resampling, mean, or zero). R--C denotes reference--candidate pairs; C--C denotes candidate--candidate pairs. Each entry is the KL misranking rate (\%), followed in parentheses by the mean held-out behavioral deficit of the preferred, worse circuit (percentage points, pp). Rates use all eligible pairs, including ties; deficits average only strict misrankings. Numeric task labels are InterpBench IDs. A dash in parentheses means no reversals; a bare dash means no eligible pairs.}\label{tab:s3kl}
\small
\setlength{\tabcolsep}{4pt}
\renewcommand{\arraystretch}{1.08}
\begin{tabular*}{\linewidth}{@{\extracolsep{\fill}}lrrrrrr@{}}
\toprule
\multicolumn{1}{l}{Task} & \multicolumn{2}{c}{Resampling} & \multicolumn{2}{c}{Mean} & \multicolumn{2}{c}{Zero}\\
\cmidrule(lr){2-3}\cmidrule(lr){4-5}\cmidrule(lr){6-7}
 & \multicolumn{1}{r}{R--C} & \multicolumn{1}{r}{C--C} & \multicolumn{1}{r}{R--C} & \multicolumn{1}{r}{C--C} & \multicolumn{1}{r}{R--C} & \multicolumn{1}{r}{C--C}\\
\midrule
\multicolumn{7}{l}{\emph{Human suite}}\\
IOI & 4.4 (0.33) & 6.1 (2.75) & 33.9 (6.60) & 24.7 (6.27) & 3.9 (1.52) & 19.5 (14.99)\\
Greater-Than & 1.7 (0.33) & 4.4 (1.21) & 1.7 (0.33) & 6.2 (1.09) & 55.6 (26.15) & 53.4 (25.52)\\
Docstring & 8.7 (0.90) & 5.6 (1.46) & 14.0 (2.56) & 12.7 (3.63) & 29.3 (4.96) & 18.6 (5.46)\\
Acronym & 5.8 (1.48) & 7.8 (4.01) & 15.8 (7.06) & 20.2 (15.70) & 5.8 (18.44) & 33.8 (25.48)\\
\midrule
\multicolumn{7}{l}{\emph{InterpBench suite}}\\
113 & 2.2 (0.05) & 1.5 (0.21) & 12.2 (0.46) & 8.2 (1.15) & 11.7 (0.37) & 8.4 (2.07)\\
97 & 0.0 (--) & 4.5 (11.63) & 0.0 (--) & 9.9 (23.02) & 0.0 (--) & 18.2 (23.15)\\
2 & 1.7 (0.07) & 0.7 (0.12) & 1.1 (0.20) & 1.1 (2.03) & 36.7 (30.80) & 25.1 (32.54)\\
82 & 0.6 (0.11) & 1.4 (5.07) & 0.6 (0.19) & 3.9 (10.70) & 9.4 (29.34) & 23.9 (33.28)\\
111 & 5.0 (0.29) & 3.0 (0.60) & 15.6 (0.95) & 7.7 (0.96) & 77.2 (5.43) & 70.2 (4.47)\\
45 & 0.0 (--) & 0.9 (1.66) & 6.1 (2.28) & 5.0 (4.56) & 11.7 (12.57) & 11.9 (19.31)\\
58 & 2.8 (0.34) & 2.3 (0.95) & 0.6 (0.41) & 2.4 (5.98) & 35.0 (54.63) & 39.9 (43.39)\\
93 & 1.1 (0.07) & 1.2 (1.71) & 0.0 (--) & 1.9 (9.88) & 71.7 (61.29) & 70.1 (54.92)\\
103 & 6.7 (0.20) & 2.9 (1.05) & 0.0 (--) & 3.5 (14.10) & 15.0 (8.99) & 20.8 (19.90)\\
25 & 9.3 (0.25) & 16.3 (2.36) & 12.7 (0.23) & 21.6 (4.34) & 37.3 (50.36) & 48.1 (44.18)\\
\bottomrule
\end{tabular*}
\end{table}

\paragraph{Misranking across intervention contexts.}
Table~\ref{tab:s3kl} reports KL misranking and the associated behavioral deficits for each task. We observe candidate--candidate reversals under ordinary resampling in all fourteen tasks: rates range from 4.4\% to 7.8\% in the Human suite and from 0.7\% to 16.3\% in InterpBench. These failures occur even though validation faithfulness and held-out behavior use the same intervention family. Thus, misranking does not require mean or zero ablation: even activations drawn from other prompts can create an evaluation context in which faithfulness prefers the worse circuit.

Changing the intervention can substantially alter this preference. On Greater-Than, the C--C misranking rate rises from 4.4\% under resampling to 53.4\% under zero replacement. Yet the effect depends on the circuits: on IOI, R--C misranking is more frequent under mean (33.9\%) than zero (3.9\%). Throughout these comparisons, we keep the candidates and their held-out behavioral ordering fixed. The ranking changes therefore arise from the evaluation intervention. This pattern is consistent with context distortion: different replacement signals change the conditions under which retained components compute and can favor different circuits. In Section~\ref{sec:repair}, we test this explanation by restoring selected intact-model signals and asking whether the wrong preference is reversed.

The reversals include cases where a candidate receives a higher faithfulness score than the known-mechanism reference despite reproducing substantially less of the model's behavior. Yet even when the score correctly favors the reference, it can still choose poorly among alternative circuits. On InterpBench task 97, we observe no reference--candidate reversal under any intervention, yet faithfulness misranks 4.5\%, 9.9\%, and 18.2\% of candidate--candidate pairs under resampling, mean, and zero, respectively. A score can therefore recognize the reference and still favor the worse circuit when choosing among alternatives.

\begin{figure}[!t]
\centering
\includegraphics[width=\linewidth]{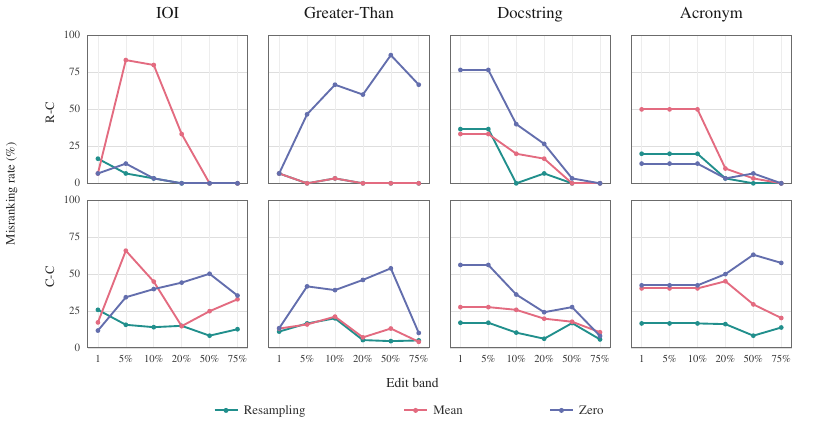}
\caption{\textbf{Misranking spans local and broad changes to a reference.} Each column is a Human task. The horizontal axis gives one component replaced (1) or the percentage of reference components replaced at fixed size. The vertical axis is the KL misranking rate over all eligible pairs, including ties. R--C (top) compares the reference with candidates; C--C (bottom) compares candidates within a band. Colors distinguish resampling, mean, and zero evaluation. Merged bands share masks; overall C--C results also include between-band pairs.}
\label{fig:bands}
\par\medskip
\begingroup
\centering
\captionof{table}{\textbf{Misranking can favor substantially worse circuits.} Candidate--candidate KL failures in the controlled experiment, grouped by suite and validation intervention. Behavioral deficit is the held-out performance lost by choosing the worse circuit; Median and IQR report its median and interquartile range in percentage points (pp). Total counts all strict KL misrankings; Deficit $\geq5$ pp counts those losing at least five points. Full distributions appear in Table~\ref{tab:s3gapdist}.}
\label{tab:severity}
\small
\setlength{\tabcolsep}{6pt}
\renewcommand{\arraystretch}{1.12}
\begin{tabular*}{\linewidth}{@{\extracolsep{\fill}}lrrrr@{}}
\toprule
& \multicolumn{2}{c}{Behavioral deficit (pp)} & \multicolumn{2}{c}{KL misrankings (count)}\\
\cmidrule(lr){2-3}\cmidrule(l){4-5}
Intervention & Median & IQR & Total & Deficit $\geq 5$ pp\\
\midrule
\multicolumn{5}{@{}l}{\textit{Human suite}}\\
Resampling & 1.00 & [0.33, 2.33] & 2,879 & 333\\
Mean & 3.00 & [1.00, 8.67] & 7,824 & 2,968\\
Zero & 14.11 & [2.33, 42.33] & 16,234 & 10,607\\
\midrule
\multicolumn{5}{@{}l}{\textit{InterpBench}}\\
Resampling & 0.52 & [0.11, 4.56] & 4,787 & 1,138\\
Mean & 1.54 & [0.33, 9.93] & 9,382 & 3,399\\
Zero & 13.67 & [3.52, 60.48] & 51,841 & 36,211\\
\bottomrule
\end{tabular*}
\endgroup

\end{figure}

\paragraph{Behavioral cost of misranking.}
Following the faithfulness ranking can lead to substantial losses in held-out performance. Under zero replacement, the preferred candidate in misranked C--C pairs has, on average, 25.52 percentage points lower $Q$ on Greater-Than and 25.48 points lower $Q$ on Acronym than the equally sized alternative. In reference--candidate comparisons on InterpBench task 93, the mean deficit reaches 61.29 points. These reversals therefore involve circuits with substantially different behavioral performance, not merely near-tied candidates. The problem also occurs under ordinary resampling: Table~\ref{tab:severity} records 333 Human and 1,138 InterpBench C--C reversals with deficits of at least 5 points. Thus, choosing the more faithful circuit can sacrifice substantial behavior even under the standard resampling protocol.

\paragraph{Misranking across edit sizes.}
We next separate candidates by how much of the reference we replace (Figure~\ref{fig:bands}). We find reversals after changing just one component, so evaluation can fail even between circuits close to the reference. Larger edits also expose candidates that reproduce much less behavior yet score above it. On Greater-Than, zero replacement misranks 26/30 reference comparisons in the 50\% band and 20/30 in the 75\% band.

We do not, however, observe a uniform increase with edit size. For Docstring, the zero R--C count falls from 23/30 after one edit to zero at 75\%. The amount of editing alone does not determine whether faithfulness makes the wrong choice. The pattern instead depends on which components the candidate retains and how that computation behaves in the evaluation context.

\paragraph{Misranking across faithfulness metrics.}
We repeat the analysis with LD and PD on the same pair pool. Across all fourteen tasks, resampling C--C rates are 3.7\% for KL, 6.1\% for LD, and 5.2\% for PD (Table~\ref{tab:metrics3}). Although the metrics do not always fail on the same pairs, we find many failures on which they agree. Across interventions, all three misrank 5,809 Human and 14,864 InterpBench C--C comparisons; 2,378 and 6,851 of these, respectively, have behavioral deficits of at least 5 pp. Thus, changing the scoring formula does not remove the problem, and agreement among all three scores does not ensure a better choice. We report the full overlap analysis in Appendix~\ref{app:overlap}.

\section{Faithfulness Can Misrank Circuits in Automated Discovery}
\label{sec:discovery}
Our controlled experiments show that faithfulness evaluation alone can misrank circuits. We now ask whether this failure also occurs in practical circuit discovery: can faithfulness favor a worse circuit among the candidates returned by automated methods? We run EAP, EAP-IG, ACDC, and Edge-SP, then compare their outputs using the same evaluation as in Section~\ref{sec:controlled}.

\paragraph{Discovered candidate pools.}
We run each method with 30 seeds per task and 100 discovery prompts per seed. We share these prompts across methods and keep them separate from evaluation. Across fourteen tasks, the four methods produce 1,680 outputs. We allow 30 minutes per run and at most 300 updates for Edge-SP, then use method-specific extraction to obtain circuits at the reference's exact executable size (Appendix~\ref{app:section4}). We reuse the validation and test sets, mean bank, and donor assignments from Section~\ref{sec:controlled}. As before, we compare distinct masks; for candidate--candidate comparisons, we also hold the discovery method fixed.

We first examine the behavior of the circuits available for selection (Table~\ref{tab:s4perf}). Their quality varies considerably. On InterpBench task 97, EAP candidates average 25.82\% $Q$, whereas EAP-IG candidates average 100\%. On Docstring, several methods produce candidates that exceed the reference's 34.00\% $Q$. We therefore cannot infer a candidate's quality from its method or its relation to the reference. We measure its held-out behavior and then ask whether faithfulness orders the available circuits accordingly.

\begin{table}[t]
\centering
\caption{\textbf{Held-out behavior of discovered circuits.} Reference reports the reference circuit\textquotesingle s test performance $Q$ (\%); each method reports mean $Q$ over its distinct masks, with their count in parentheses. $Q$ is intact-model answer agreement under ordinary resampling, except Greater-Than\textquotesingle s semantic completion accuracy. Numeric task labels are InterpBench IDs. All masks have the task\textquotesingle s reference size.}\label{tab:s4perf}
\small
\setlength{\tabcolsep}{4pt}
\renewcommand{\arraystretch}{1.08}
\begin{tabular*}{\linewidth}{@{\extracolsep{\fill}}lrrrrr@{}}
\toprule
Task & Reference & EAP & EAP-IG & ACDC & Edge-SP\\
\midrule
\multicolumn{6}{l}{\emph{Human suite}}\\
IOI & 99.00 & 90.47 (30) & 98.07 (30) & 94.72 (30) & 85.58 (30)\\
Greater-Than & 99.67 & 95.79 (30) & 99.70 (30) & 98.69 (30) & 99.79 (30)\\
Docstring & 34.00 & 27.93 (24) & 34.50 (28) & 35.99 (24) & 37.19 (30)\\
Acronym & 87.67 & 84.26 (6) & 83.00 (4) & 85.50 (6) & 33.12 (30)\\
\midrule
\multicolumn{6}{l}{\emph{InterpBench suite}}\\
113 & 99.93 & 59.07 (30) & 100.00 (30) & 97.80 (30) & 97.45 (30)\\
97 & 100.00 & 25.82 (30) & 100.00 (30) & 97.19 (5) & 99.97 (30)\\
2 & 97.78 & 99.70 (30) & 100.00 (30) & 98.46 (16) & 99.38 (30)\\
82 & 99.44 & 98.07 (30) & 100.00 (30) & 96.62 (30) & 95.96 (30)\\
111 & 97.33 & 97.07 (30) & 99.79 (30) & 97.83 (30) & 99.49 (30)\\
45 & 98.22 & 98.01 (30) & 99.70 (30) & 97.84 (29) & 93.85 (30)\\
58 & 98.78 & 95.69 (30) & 100.00 (25) & 98.13 (15) & 99.94 (30)\\
93 & 99.89 & 97.57 (30) & 100.00 (30) & 98.01 (12) & 96.84 (30)\\
103 & 99.41 & 96.66 (30) & 100.00 (30) & 97.19 (22) & 99.81 (30)\\
25 & 99.33 & 99.14 (27) & 99.97 (30) & 99.43 (6) & 99.54 (30)\\
\bottomrule
\end{tabular*}
\end{table}

\begin{table}[t]
\centering
\caption{\textbf{Faithfulness misranks actual discovery outputs.} Results are grouped by discovery method and validation intervention. R--C denotes reference--candidate pairs; C--C denotes candidate--candidate pairs. Each entry gives the KL misranking rate (\%) and, in parentheses, the mean held-out behavioral deficit among strict reversals (percentage points, pp). We compare equal-size circuits within each task and method. Rate denominators include behavioral and score ties. A dash in parentheses means no reversals; a bare dash means no eligible pairs.}
\label{tab:discovery}\label{tab:s4Human4}\label{tab:s4Interp10}
\small
\setlength{\tabcolsep}{4pt}
\renewcommand{\arraystretch}{1.12}
\begin{tabular*}{\linewidth}{@{\extracolsep{\fill}}lrrrrrr@{}}
\toprule
Method & \multicolumn{2}{c}{Resampling} & \multicolumn{2}{c}{Mean} & \multicolumn{2}{c}{Zero}\\
\cmidrule(lr){2-3}\cmidrule(lr){4-5}\cmidrule(lr){6-7}
 & R--C & C--C & R--C & C--C & R--C & C--C\\
\midrule
\multicolumn{7}{@{}l}{\emph{Human suite}}\\
EAP & 5.6 (2.07) & 15.9 (1.82) & 38.9 (7.42) & 23.1 (2.59) & 27.8 (4.92) & 44.1 (4.07)\\
EAP-IG & 42.4 (1.51) & 24.4 (1.10) & 56.5 (1.23) & 46.2 (1.54) & 28.3 (1.14) & 37.6 (1.44)\\
ACDC & 11.1 (1.63) & 20.6 (0.87) & 52.2 (3.60) & 25.2 (1.26) & 52.2 (1.61) & 39.9 (1.40)\\
Edge-SP & 15.0 (0.33) & 13.2 (1.96) & 46.7 (5.28) & 18.8 (4.25) & 15.8 (8.43) & 42.9 (11.75)\\
\midrule
\multicolumn{7}{@{}l}{\emph{InterpBench suite}}\\
EAP & 7.4 (0.48) & 7.2 (0.32) & 11.8 (0.58) & 10.8 (1.04) & 47.8 (1.73) & 29.5 (6.00)\\
EAP-IG & 0.3 (0.15) & 2.5 (0.11) & 8.1 (1.16) & 2.4 (0.23) & 44.7 (1.42) & 5.7 (0.26)\\
ACDC & 10.8 (0.54) & 8.4 (0.34) & 17.9 (0.71) & 25.6 (1.09) & 40.0 (1.57) & 51.9 (1.52)\\
Edge-SP & 2.0 (0.36) & 16.2 (0.12) & 4.0 (0.58) & 15.4 (0.30) & 42.3 (1.31) & 38.9 (0.31)\\
\bottomrule
\end{tabular*}
\end{table}

\begin{table}[t]
\centering
\caption{\textbf{Three scores can agree on the worse candidate.} Source distinguishes controlled edits from discovery outputs; Suite identifies the evaluated tasks. Evaluated pairs counts candidate--candidate pair--intervention comparisons, including ties; the same circuit pair evaluated under different interventions contributes separately. All-three failures counts comparisons where KL, LD, and PD all prefer the circuit with lower held-out $Q$. Deficit $\geq5$ pp counts the subset of these all-three failures with a behavioral deficit of at least five percentage points.}
\label{tab:consensus}
\small
\setlength{\tabcolsep}{5pt}
\renewcommand{\arraystretch}{1.16}
\begin{tabular*}{\linewidth}{@{\extracolsep{\fill}}llrrr@{}}
\toprule
Source & Suite & Evaluated pairs & All-three failures & Deficit $\geq5$ pp\\
\midrule
Controlled edits & Human & 151,605 & 5,809 & 2,378\\
Controlled edits & InterpBench & 467,958 & 14,864 & 6,851\\
\midrule
Discovery outputs & Human & 15,948 & 1,354 & 144\\
Discovery outputs & InterpBench & 45,267 & 3,019 & 150\\
\bottomrule
\end{tabular*}
\end{table}

\paragraph{Faithfulness can select worse discovery outputs.}
Misranking persists when candidates come from automated discovery. Under ordinary resampling, KL favors the worse circuit in 15.9\% of Human EAP candidate pairs, 24.4\% for EAP-IG, 20.6\% for ACDC, and 13.2\% for Edge-SP (Table~\ref{tab:discovery}). We also observe reversals in InterpBench, with rates ranging from 2.5\% for EAP-IG to 16.2\% for Edge-SP. These comparisons hold both discovery method and circuit size fixed. Thus, the failure is not simply that a method fails to find a good circuit: it has already produced a better alternative, but faithfulness gives us a reason to reject it.

The evaluation intervention can further change which outputs appear preferable. Within the same Human EAP-IG pool, the C--C misranking rate rises from 24.4\% under resampling to 46.2\% under mean replacement. For InterpBench ACDC, it rises from 8.4\% under resampling to 51.9\% under zero. These changes affect more than the frequency of reversals. Under zero replacement, following faithfulness among misranked Human Edge-SP pairs loses an average of 11.75 percentage points of held-out performance. The candidates and their behavioral ordering remain fixed; changing their evaluation context changes how often, and how severely, the score chooses poorly. This extends Section~\ref{sec:controlled}'s finding to actual discovery outputs: the intervention used to judge a circuit can work against the recovery goal.

\paragraph{Misranking beyond semi-synthetic benchmarks.}
InterpBench provides known computations, but its semi-synthetic models do not capture the full complexity of mechanisms learned through language-model pretraining. The Human suite therefore provides a more direct test of whether misranking affects circuit analysis in pretrained language models.

We find that the problem is more pronounced in this setting. Pooling common C--C pairs across methods, Human KL misranking rates are 18.0\%, 27.6\%, and 41.2\% under resampling, mean, and zero, compared with 8.7\%, 12.0\%, and 28.9\% in InterpBench (Table~\ref{tab:metrics4}). Mean behavioral deficits are also larger in the Human suite under every intervention. Although these suites differ in tasks, models, and candidate pools, the practical message is clear: misranking is not confined to constructed benchmarks with known mechanisms. It also affects the pretrained-model circuits that automated discovery aims to explain, where following the score can incur larger behavioral losses.

\paragraph{Misranking across faithfulness metrics.}
Misranking is not specific to KL: under ordinary resampling, LD and PD misrank 16.4\% and 17.6\% of Human C--C pairs, respectively. Moreover, all three metrics share the same wrong preference in 1,354 Human and 3,019 InterpBench pair--intervention comparisons among discovery outputs, including cases with substantial behavioral losses (Table~\ref{tab:consensus}). These shared failures show that misranking cannot be dismissed as an artifact of our chosen metric. It is a recurring problem in intervention-based circuit evaluation, including when the candidates come from actual discovery methods. Replacing one faithfulness metric with another therefore does not, by itself, resolve the evaluation problem.

\begin{algorithm2e}[!b]
\caption{Partial Context Restoration}
\label{alg:context-restoration}
\small
\DontPrintSemicolon
\SetNlSkip{0pt}
\KwIn{Misranked pair $(B,W)$ with $Q(B)>Q(W)$; original intervention $a$; eligible receiving-node groups $\mathcal{G}$; validation split $V$ and held-out split $T$}
\KwOut{Confirmed partial restorations $\mathcal{C}$; the case is repaired if $\mathcal{C}\neq\emptyset$}
$P\leftarrow\mathcal{G}$; $\mathcal{V}\leftarrow\emptyset$\;
\For{$f\in(0.8,0.4,0.2,0.1,0.05)$}{
  $k\leftarrow\max(1,\operatorname{round}(f|\mathcal{G}|))$; $P_{\mathrm{next}}\leftarrow\mathcal{G}$\;
  \For{$j=1,\ldots,10$}{
    Sample $R\subseteq P$ with $|R|=k$ using the predetermined draw\;
    \If{$\operatorname{Repairs}(V,R)$}{
      \If{this is the first validation success at this level}{
        $P_{\mathrm{next}}\leftarrow R$\;
      }
      $\mathcal{V}\leftarrow\mathcal{V}\cup\{R\}$\;
    }
  }
  $P\leftarrow P_{\mathrm{next}}$\;
}
$\mathcal{C}\leftarrow\{R\in\mathcal{V}:\operatorname{Repairs}(T,R)\}$\;
\Return{$\mathcal{C}$}\;
\end{algorithm2e}

\section{Context Distortion Contributes to Misrankings}
\label{sec:repair}
Why does faithfulness sometimes favor the worse circuit? To evaluate a circuit, we replace signals outside it, but the retained components still depend on those signals. Two circuits can respond differently to the replacements, so the new context may favor one that reproduces the model's original behavior less well. We test this \emph{context distortion} explanation by selectively restoring intact-model signals. If these replacements contribute to the wrong preference, undoing some of them should allow faithfulness to recognize the originally better circuit.

\paragraph{Misranking cohort selection and context restoration.}
Searching for context restorations across every misranked pair would be computationally expensive. We therefore select 100 distinct equal-size pairs from Section~\ref{sec:discovery}, reusing its validation and held-out prompts. We use validation prompts to find restorations and held-out prompts to test whether those restorations generalize to new inputs. If restoring the same signal locations reverses the wrong preference on new inputs, this supports context distortion as a cause of the ranking failure, rather than a repair fitted only to the search examples.

We require a behavioral deficit of at least 0.5 pp (mean 3.36 pp) and strict KL misranking on both splits. These criteria exclude near-ties below the threshold and failures that disappear on held-out evaluation. We allocate 50 cases to each suite and include 50 resampling, 25 mean, and 25 zero failures. The selected cases cover all fourteen tasks and all 51 task--method combinations with eligible cases. Table~\ref{tab:s5allocation} gives the allocation across tasks.

For each recipient prompt $x$, we cache the intact-model signals $h^M_e(x)$. We then choose a subset $R$ of excluded connections and restore their recipient-intact signals in place of the original intervention. Writing $r_{a,e}(x)$ for the original donor, mean, or zero replacement, we supply
\begin{equation}
\widetilde h_e(x;S,a,R)=
\begin{cases}
 h_e^{S,a,R}(x), & e\in S \quad\text{(computed in the current forward pass)},\\
 h_e^M(x), & e\notin S,\ e\in R \quad\text{(recipient-intact context)},\\
 r_{a,e}(x), & e\notin S,\ e\notin R \quad\text{(original replacement)}.
\end{cases}
\label{eq:restoration}
\end{equation}
We restore the selected signals wherever each circuit excludes them, keeping circuit masks and all other replacements fixed. We then test whether KL favors the circuit with higher original held-out $Q$.

Our goal is to test whether context distortion contributes to misranking: can restoring selected intact-model signals reverse the wrong preference while the circuit structures remain fixed? We therefore use a simple, bounded search for sufficient partial restorations, rather than developing a new discovery method or finding the smallest repair. Algorithm~\ref{alg:context-restoration} summarizes the procedure. 

We group excluded incoming connections by their receiving node and denote the eligible group set by $\mathcal{G}$. The search tries ten subsets at each of five levels, measured relative to $|\mathcal{G}|$, for at most 50 validation attempts per case. In the algorithm, $\operatorname{Repairs}(D,R)$ means that restoring the connections in groups $R$ using Equation~\ref{eq:restoration} makes KL prefer the originally better circuit $B$ over $W$ on split $D$, beyond the predefined numerical tolerance (Appendix~\ref{app:repair}). Circuit masks, the original $Q$ ordering, and replacements elsewhere remain fixed.

\begin{figure}[!t]
\centering
\begin{subfigure}[t]{0.428\linewidth}
\centering
\includegraphics[width=\linewidth]{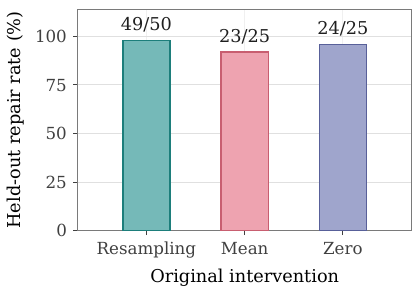}
\caption{Repair across interventions.}
\label{fig:repairs-interventions}
\end{subfigure}\hfill
\begin{subfigure}[t]{0.55\linewidth}
\centering
\includegraphics[width=\linewidth]{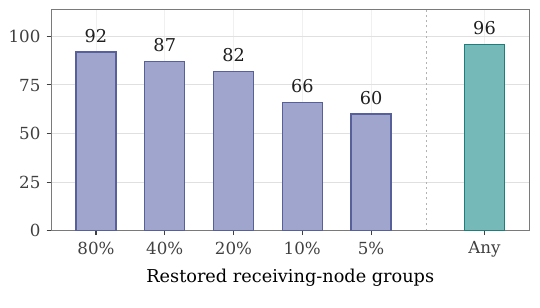}
\caption{Repair with partial restoration.}
\label{fig:repairs-levels}
\end{subfigure}
\caption{\textbf{Restoring context reverses persistent KL misrankings.} Bar heights show confirmed repair rates (\%): a subset must correct both validation and held-out rankings. (a) Labels give repaired/selected cases for each original intervention, counting a case once across levels. (b) The horizontal axis gives the fraction of eligible receiving-node groups restored; labels count confirmed cases out of 100. Each level receives ten attempts. Any is the union across levels (96/100); level counts overlap. A group contains excluded incoming connections to one receiving node.}
\label{fig:repairs}
\end{figure}

\begin{table}[!b]
\centering
\caption{\textbf{Partial context restoration transfers beyond validation.} Nominal level is the percentage of eligible receiving-node groups restored. Validation counts cases with a ranking repair among ten attempts; Held-out confirmed counts cases where the same subset also repairs the held-out ranking, each out of 100 cases. Groups restored and Sites restored are median numbers of receiving-node groups and excluded connections in the first confirmed subset per case at that level. Cases may succeed at multiple levels.}
\label{tab:repairsize}
\small
\setlength{\tabcolsep}{4pt}
\begin{tabular*}{\linewidth}{@{\extracolsep{\fill}}lrrrr@{}}
\toprule
Nominal level & Validation & Held-out confirmed & Groups restored & Sites restored\\
\midrule
80\% & 92 / 100 & 92 / 100 & 10 & 283\\
40\% & 88 / 100 & 87 / 100 & 5 & 128\\
20\% & 84 / 100 & 82 / 100 & 2 & 83.5\\
10\% & 76 / 100 & 66 / 100 & 1 & 31\\
5\% & 72 / 100 & 60 / 100 & 1 & 17\\
\bottomrule
\end{tabular*}
\end{table}

\paragraph{Context restoration reverses the wrong preference.}
Restoring selected intact-model signals repairs 96 of 100 persistent misrankings: 49/50 Human cases and 47/50 InterpBench cases. The repairs span ordinary resampling (49/50), mean replacement (23/25), and zero replacement (24/25), as well as reference--candidate (46/47) and candidate--candidate comparisons (50/53) (Figure~\ref{fig:repairs}).

The significance is that we obtain these reversals without finding a new circuit or changing which circuit originally performed better. We change only selected signals supplied by the evaluation context, and faithfulness then recognizes the better circuit on both validation and held-out prompts. This directly supports context distortion as a cause of the wrong preference in these cases. The 49/50 repairs under resampling are particularly revealing: using activations from real prompts does not ensure that those signals provide an appropriate context for the retained computation. A donor signal can be natural in its source prompt yet disrupt the circuit when supplied to a different recipient.

\paragraph{Small context changes can determine circuit preference.}
A broad restoration is not always necessary. We confirm repairs in 60 cases at the nominal 5\% level; among the first confirmed subsets at this level, the median restoration contains one receiving-node group and 17 excluded sites (Table~\ref{tab:repairsize}). Most of the original intervention remains in place, yet restoring this small part of the context is sufficient to reverse the ranking.

These small repairs suggest that a circuit's behavior can depend strongly on particular incoming pathways, rather than on every part of its surrounding context. Such pathways may supply task-relevant information, but they may also play a supporting role: modulating or stabilizing the conditions under which the retained computation works. Replacing their signals can therefore change which circuit appears more faithful, even when most of the intervention remains unchanged.

This suggests a direction for circuit discovery: identifying the components that perform a task may be insufficient without also accounting for the contextual inputs on which they depend. Future methods could explicitly represent these dependencies and test whether their source components belong in the recovered circuit. Our results motivate this distinction between the computation a circuit performs and the supporting context it requires.

\section{Related Work}
\label{sec:related}
\paragraph{Circuit discovery and recovery benchmarks.}
Automated methods have made it easier to find circuit candidates. ACDC prunes connections through activation patching \citep{conmy}; EAP estimates edge effects with gradients \citep{syed}, and EAP-IG uses integrated gradients \citep{hanna,ig}. Edge-SP learns an edge mask through subnetwork optimization \citep{conmy,movement}. Other work scales sparse edge-mask optimization \citep{edgepruning} or searches over interpretable features \citep{marks}. But finding candidates is only part of recovery: the score used to choose among them must recognize the better circuit. We test that choice among outputs of ACDC, EAP, EAP-IG, and Edge-SP. We also use known computations to study the score before search enters. Tracr compiles programs into transformers \citep{tracr}; InterpBench supplies semi-synthetic models with specified circuits \citep{gupta}, while MIB standardizes evaluations of circuit and causal-variable localization \citep{mib}. Alongside circuits from published analyses, the specified InterpBench circuits let us construct controlled comparisons in which discovery plays no role.

\paragraph{Reliability of circuit evaluation.}
Earlier work shows that the way we intervene matters. Ablation choices change faithfulness measurements \citep{miller}, and corruption and scoring choices affect activation-patching conclusions \citep{zhang,patchguide}. Adversarial Circuit Evaluation finds inputs on which published circuits diverge from the full model \citep{uit}, while statistical circuit tests examine preservation, localization, and minimality \citep{shi}. Subspace-patching studies show that an intervention can alter outputs through a dormant pathway \citep{illusion}. We ask a different question that follows from these concerns: can the score choose a worse circuit when both candidates are available and matched in size? We show that it can. We trace the wrong preference to context distortion and test it by restoring selected intact-model signals. This account may also help explain why faithfulness changes with ablation choices: different replacements create different environments for the retained computation.

\section{Discussion and conclusion}
We show that better search alone cannot guarantee circuit recovery, because faithfulness can prefer a worse candidate even when a better one is available. We first expose this failure with controlled edits to equally sized circuits, without involving a discovery algorithm. In the most striking cases, faithfulness even scores a behaviorally worse candidate above a benchmark's known-mechanism reference circuit. We then find the same misranking among circuits produced by EAP, EAP-IG, ACDC, and Edge-SP, where it can misdirect real selection. Misranking also occurs under ordinary resampling. Among discovery outputs in the Human suite, its frequency is broadly similar across KL, logit-difference, and probability-difference scores.

We trace the wrong preference to context distortion. To isolate a circuit's effect, we replace signals outside it, but that intervention changes the computational environment in which the circuit operates. We test this explanation by restoring selected signals from the intact model. This reverses 96 of 100 selected persistent misrankings without changing the circuit structures. In these cases, the context created by evaluation made faithfulness favor the circuit that reproduced the original model's behavior less well.

\paragraph{Implications for discovery and evaluation.}
Our findings expose a dilemma at the heart of intervention-based circuit analysis. We use interventions to isolate a circuit's effect, yet replacing excluded signals can change the computational environment in which the circuit operates and make a worse candidate appear more faithful. When that score guides search, context distortion can steer discovery toward the wrong circuit. When it is used to judge a final result, it can endorse the wrong explanation. These findings call for rethinking both how circuit discovery is performed and how its results are evaluated.

\paragraph{Limitations.}
Our behavioral criterion $Q$ measures agreement with the intact model under ordinary resampling. A circuit can match those outputs without recovering the actual mechanism, so we establish a ranking failure relative to behavior, not mechanism identity. Context restoration is a diagnostic, not a replacement faithfulness score: restored signals may supply information the circuit would not compute on its own. We use it to test whether changing the intervention context reverses a wrong preference. How to isolate a circuit's effect without creating that distortion remains open.

\pagebreak
\bibliography{iclr2027_conference}
\bibliographystyle{iclr2027_conference}

\clearpage
\appendix
\makeatletter
\setlength{\@fptop}{0pt}
\setlength{\@fpsep}{18pt}
\makeatother

\section{Shared experimental and analysis details}
\label{app:protocol}\label{app:provenance}
\subsection{Models, hardware, and evaluation prompts}
We summarize the tasks and their reference sources in Table~\ref{tab:tasks}. For IOI, Greater-Than, and Acronym, we use GPT-2 small. For Docstring, we use the four-layer attention-only model of width 512 studied with its reference circuit. We represent IOI, Greater-Than, and Docstring circuits with factorized attention/residual edges, and Acronym circuits with attention heads. Each InterpBench task supplies its own compiled transformer and native ground-truth computation. We match size within these representations; a head in one task and an edge in another are not interchangeable units.

We run the experiments on Trillium H100 GPU nodes in a fixed container environment, using 30 seeds (0--19 and 810--819). We allow up to 30 minutes per instance in Sections 3--5. Fixed scoring finishes within this limit, while discovery and restoration also follow their algorithmic stopping rules. To compare controlled and discovered candidates under the same conditions, we reuse the 100 validation prompts, 100 held-out prompts, 100-prompt mean bank, and three donor assignments in Sections 3 and 4. We keep validation and held-out prompt identities disjoint and give discovery separate prompts.

We preserve the prompt strata used for each Human task: the saved ABBA template (a fixed ordering of the two name roles) for IOI, century strata for Greater-Than, the fixed 41-token stratum for Docstring, and four-word prompts for Acronym. For InterpBench, we retain the saved prompt order and valid-position masks. We average the three donor draws within each recipient prompt before averaging across prompts, so the three draws contribute to one prompt-level measurement. For mean ablation, we reuse the frozen mean bank; for zero ablation, we replace the same excluded signals with zero.

\subsection{Scores and behavioral targets}
We compute KL between the intact and intervened output distributions at the scored positions. For Human tasks, we use the original vocabulary distribution; for Acronym, we score the three teacher-forced letter positions, conditioning each prediction on the preceding reference letters. For InterpBench, we use the original categorical output space and the nine valid positions per prompt. We first average position-level contributions within each prompt and then average across prompts.

For the additional metrics, let $P$ contain the task-correct classes and $N$ the remaining valid task classes. Let $z_j$ be the intervened circuit's logit for class $j$, and $p_j$ its probability under the full softmax. At each output position, we compute
\begin{equation}
 F^{\mathrm{LD}}=\frac{1}{|P|}\sum_{j\in P}z_j-\frac{1}{|N|}\sum_{j\in N}z_j,
 \qquad
 F^{\mathrm{PD}}=\sum_{j\in P}p_j-\sum_{j\in N}p_j.
\label{eq:margins}
\end{equation}
We then use the same position and donor averaging as for KL. This keeps aggregation fixed when we compare scoring metrics.

We define $P$ and $N$ from the dataset rather than from candidate predictions. For IOI and Docstring, $P$ is the correct answer option and $N$ contains the other fixed options. For Acronym, we use the correct letter token and the other 25 alphabet tokens at each teacher-forced position. For Greater-Than, $P$ contains the valid two-digit year endings strictly above the prompt's starting ending, and $N$ contains the remaining valid endings. For InterpBench, we use the dataset-correct class and the other original classes. In task order 113, 97, 2, 82, 111, 45, 58, 93, 103, and 25, these tasks have 30, 64, 26, 16, 11, 16, 11, 11, 11, and 56 classes. We take PD probabilities from the full original softmax without renormalizing over the answer options.

For the ordinary agreement criterion, we compare each circuit's task-native answer with the intact model's answer under ordinary resampling, using the same three donor draws and prompt aggregation. For Greater-Than, we instead measure whether the predicted two-digit ending strictly exceeds the starting ending. The reference reaches 99.67\% on this semantic criterion; we do not replace it with exact-token agreement. In every task, we keep Q fixed when comparing KL, LD, and PD or changing the validation intervention.

\subsection{Total-pair denominators, ties, and uncertainty}
We deduplicate circuit masks within each task and candidate pool before forming comparisons. We include each reference--candidate identity once and each unordered pair of distinct candidates once, requiring equal actual executable size for the primary analysis. We divide the number of strict reversals by the total pair count, including Q ties and score ties. A tie does not count as misranking, but it also does not show that the score can distinguish the circuits. When deduplication leaves no available pairs, we report the cell as unavailable rather than assigning it a zero rate. For cross-metric comparisons, we use only the exact pair identities available for all three scores.

We use the same scale-aware tolerance throughout: values $a$ and $b$ are tied when $|a-b|\leq10^{-6}+10^{-6}\max(|a|,|b|)$. We apply this rule to both Q and faithfulness before rounding values for display. When we report a mean Q deficit alongside a misranking rate, we average over the strict reversals in that cell. We impose no minimum behavioral gap in Sections 3 or 4; the 0.5-pp threshold applies only when selecting the Section 5 restoration cohort.

We estimate paired bootstrap intervals by resampling prompts, keeping the two circuits' contributions paired and averaging donors within each prompt. We use 10,000 replicates wherever we report these intervals and retain the semantic Q criterion for Greater-Than. Because a circuit can recur across pairs and interventions, we use pooled counts and deficit distributions to describe the evaluated candidate pools rather than treating every comparison as an independent observation.

\section{Section 3: controlled candidates, reference conversion, and additional results}
\label{app:section3}
\subsection{Converting InterpBench references without dropping their computation}
We take InterpBench's native node computation as the reference to preserve. Simply projecting its listed graph edges into our factorized edge evaluator can omit outgoing contributions from retained nodes. The nominal node set may look unchanged, but its execution is not. We therefore construct an outgoing-edge closure that preserves the contributions made by the native reference.

We map each retained attention output from the native head-output hook to its residual-stream result, and each retained MLP postactivation to its MLP output. We keep the input embedding as the root source. For every retained native source, we then retain all outgoing connections in the factorized Q/K/V and residual executor, while intervening on complementary sources. This supplies the retained source's contribution everywhere the native node execution supplies it. The resulting circuit can contain many more edges than a direct projection: for task 113, the 30 projected edges become a 587-edge executable closure.

We check this conversion against native reference execution under resampling, mean, and zero on both frozen 100-prompt splits, using three donor draws for resampling. We include only validated conversions; this leaves out \texttt{ioi\_next\_token}, whose reference closure remains unresolved. For the ten included tasks in Table~\ref{tab:conversion}, we obtain the same native and converted reference answers at every scored position, with a maximum absolute logit discrepancy below $8\times10^{-13}$. These checks establish that the two representations execute equivalently on the evaluated prompts. We separately measure how well that reference matches the intact model through Q. We use the validated closures in both the controlled and discovery experiments.
\begin{table}[t]
\centering
\caption{\textbf{InterpBench reference conversion preserves native execution.} Task is the benchmark ID. Projected edges counts the direct projection of the listed reference graph; Closure edges counts all outgoing connections retained to reproduce native node execution. Max logit error is the largest absolute output-logit discrepancy; Max RMS error is the largest root-mean-square discrepancy between native and converted outputs. Maxima cover both 100-prompt splits and resampling, mean, and zero conditions. Their answers agree at every scored position.}\label{tab:conversion}
\footnotesize
\setlength{\tabcolsep}{3pt}
\begin{tabular}{lrrrr}
\toprule
Task & Projected edges & Closure edges & Max logit error & Max RMS error\\
\midrule
113 & 30 & 587 & 3.02e-14 & 2.56e-15\\
97 & 20 & 127 & 1.39e-13 & 1.40e-14\\
2 & 14 & 177 & 7.19e-13 & 1.37e-14\\
82 & 10 & 176 & 9.95e-14 & 1.28e-14\\
111 & 14 & 112 & 7.67e-13 & 3.24e-14\\
45 & 9 & 110 & 5.68e-14 & 8.26e-15\\
58 & 13 & 111 & 3.02e-13 & 1.08e-14\\
93 & 13 & 111 & 4.36e-13 & 1.38e-14\\
103 & 13 & 111 & 5.31e-13 & 1.34e-14\\
25 & 10 & 72 & 5.68e-13 & 8.27e-14\\
\bottomrule
\end{tabular}
\end{table}

\subsection{Size-preserving candidate generation}
We select the ten largest validated executable references: tasks 113, 97, 2, 82, 111, 45, 58, 93, 103, and 25. For a reference with $K$ components, we request either one swap or $k=\max(1,\lfloor fK+0.5\rfloor)$ swaps for $f\in\{0.05,0.10,0.20,0.50,0.75\}$. We remove $k$ reference components and add $k$ components from outside it, keeping the circuit at size $K$. Thus, a 20\% edit changes which components are retained; it does not make the circuit 20\% smaller. We use seeds 0--19 and 810--819 for each attainable distinct swap count, with the InterpBench generator offsetting each seed by $100000k$.

We merge identical masks and share masks between edit bands that round to the same $k$. This explains the repeated points for Docstring and Acronym in the band plots. We can generate a band only when both the reference and its complement contain at least $k$ components. Task 25 consequently has 150 candidates: its 54-swap, 75\% band is unattainable. For task 97, deduplication leaves 29 distinct one-edit masks rather than 30. We report the realized counts for every task in Table~\ref{tab:bandinventory}.
\begin{table*}[t]
\centering
\caption{\textbf{Controlled candidate pools by edit band.} Reference size counts executable edges, or heads for Acronym. A band entry $k\ (n)$ gives the number of reference components replaced ($k$) and distinct candidate masks ($n$); replacement preserves total size. Percentage columns specify the fraction of the reference replaced, rounded to an integer. Merged bands share masks. A dash marks a band without a separately reported pool; task 25\textquotesingle s 75\% band is unattainable.}\label{tab:bandinventory}
\scriptsize
\setlength{\tabcolsep}{3pt}
\begin{tabular}{lrrrrrrr}
\toprule
Task & Reference size & 1 edit & 5\% & 10\% & 20\% & 50\% & 75\%\\
\midrule
IOI & 963 & 1 (30) & 48 (30) & 96 (30) & 193 (30) & 482 (30) & 722 (30)\\
Greater-Than & 235 & 1 (30) & 12 (30) & 24 (30) & 47 (30) & 118 (30) & 176 (30)\\
Docstring & 24 & 1 (30) & 1 (30) & 2 (30) & 5 (30) & 12 (30) & 18 (30)\\
Acronym & 8 & 1 (30) & -- & 1 (30) & 2 (30) & 4 (30) & 6 (30)\\
113 & 587 & 1 (30) & 29 (30) & 59 (30) & 117 (30) & 294 (30) & 440 (30)\\
97 & 127 & 1 (29) & 6 (30) & 13 (30) & 25 (30) & 64 (30) & 95 (30)\\
2 & 177 & 1 (30) & 9 (30) & 18 (30) & 35 (30) & 89 (30) & 133 (30)\\
82 & 176 & 1 (30) & 9 (30) & 18 (30) & 35 (30) & 88 (30) & 132 (30)\\
111 & 112 & 1 (30) & 6 (30) & 11 (30) & 22 (30) & 56 (30) & 84 (30)\\
45 & 110 & 1 (30) & 6 (30) & 11 (30) & 22 (30) & 55 (30) & 83 (30)\\
58 & 111 & 1 (30) & 6 (30) & 11 (30) & 22 (30) & 56 (30) & 83 (30)\\
93 & 111 & 1 (30) & 6 (30) & 11 (30) & 22 (30) & 56 (30) & 83 (30)\\
103 & 111 & 1 (30) & 6 (30) & 11 (30) & 22 (30) & 56 (30) & 83 (30)\\
25 & 72 & 1 (30) & 4 (30) & 7 (30) & 14 (30) & 36 (30) & --\\
\bottomrule
\end{tabular}
\end{table*}

\subsection{Reference and edited-candidate behavioral performance}
\label{app:performance}
We report circuit behavior separately from ranking reliability in Tables~\ref{tab:perf3Human4} and \ref{tab:perf3Interp10}. Within each band, we average Q over unique candidate masks, so a circuit does not receive extra weight because it participates in many pairs. Merged bands contain the same masks and therefore have the same mean. We evaluate the reference with the same ordinary-resampling held-out criterion.

We find that larger edits generally reduce mean Q, although the amount of degradation varies by task. The Human references also begin at different levels of Q. This distinction helps interpret the ranking results: low absolute agreement tells us that a circuit preserves little behavior, while misranking tells us that the score prefers it despite a better available alternative.
\begin{table*}[t]
\centering
\caption{\textbf{Held-out behavior of Human controlled candidates.} Reference gives the original circuit\textquotesingle s $Q$ (\%); each edit-band entry gives mean candidate $Q$ with the number of distinct masks in parentheses. $Q$ is intact-model answer agreement under ordinary resampling, except Greater-Than\textquotesingle s semantic completion accuracy. Band percentages specify reference components replaced, not a reduction in circuit size. Merged bands share masks; a dash marks a band without a separately reported pool.}\label{tab:perf3Human4}
\scriptsize
\setlength{\tabcolsep}{3pt}
\begin{tabular}{lrrrrrrr}
\toprule
Task & Reference & 1 edit & 5\% & 10\% & 20\% & 50\% & 75\%\\
\midrule
IOI & 99.00 & 98.96 (30) & 94.68 (30) & 86.98 (30) & 66.16 (30) & 53.02 (30) & 47.73 (30)\\
Greater-Than & 99.67 & 99.32 (30) & 95.72 (30) & 93.39 (30) & 78.49 (30) & 55.73 (30) & 53.73 (30)\\
Docstring & 34.00 & 31.49 (30) & 31.49 (30) & 27.84 (30) & 23.77 (30) & 13.36 (30) & 10.87 (30)\\
Acronym & 87.67 & 79.29 (30) & -- & 79.29 (30) & 69.54 (30) & 48.84 (30) & 26.97 (30)\\
\bottomrule
\end{tabular}
\end{table*}

\begin{table}[!tbp]
\centering
\caption{\textbf{Held-out behavior of InterpBench controlled candidates.} Task denotes the benchmark ID. Reference gives the original circuit\textquotesingle s $Q$ (\%); each edit-band entry gives mean candidate $Q$ with the number of distinct masks in parentheses. $Q$ is intact-model answer agreement under ordinary resampling. Band percentages specify reference edges replaced while preserving size. The dash marks task 25\textquotesingle s unattainable 75\% band.}\label{tab:perf3Interp10}
\scriptsize
\setlength{\tabcolsep}{3pt}
\begin{tabular}{lrrrrrrr}
\toprule
Task & Reference & 1 edit & 5\% & 10\% & 20\% & 50\% & 75\%\\
\midrule
113 & 99.93 & 99.92 (30) & 97.22 (30) & 90.98 (30) & 58.26 (30) & 3.40 (30) & 3.11 (30)\\
97 & 100.00 & 97.25 (29) & 74.74 (30) & 54.43 (30) & 29.11 (30) & 15.91 (30) & 7.85 (30)\\
2 & 97.78 & 94.02 (30) & 77.84 (30) & 59.53 (30) & 51.93 (30) & 4.03 (30) & 3.48 (30)\\
82 & 99.44 & 98.30 (30) & 68.56 (30) & 69.28 (30) & 39.58 (30) & 12.59 (30) & 10.15 (30)\\
111 & 97.33 & 96.73 (30) & 95.13 (30) & 93.16 (30) & 92.48 (30) & 90.33 (30) & 90.19 (30)\\
45 & 98.22 & 96.77 (30) & 74.37 (30) & 55.23 (30) & 47.87 (30) & 11.83 (30) & 10.28 (30)\\
58 & 98.78 & 92.16 (30) & 84.89 (30) & 62.75 (30) & 39.68 (30) & 14.98 (30) & 10.06 (30)\\
93 & 99.89 & 92.88 (30) & 85.71 (30) & 65.47 (30) & 46.46 (30) & 15.19 (30) & 10.78 (30)\\
103 & 99.41 & 91.11 (30) & 81.37 (30) & 56.99 (30) & 42.20 (30) & 14.15 (30) & 10.14 (30)\\
25 & 99.33 & 96.72 (30) & 77.41 (30) & 57.87 (30) & 36.43 (30) & 21.23 (30) & --\\
\bottomrule
\end{tabular}
\end{table}

\FloatBarrier

\subsection{Task-level, within-band, and margin-score results}
\label{app:taskdetail}
We give the exact KL reversal counts and total pair counts in Table~\ref{tab:s3detail}. In Figure~\ref{fig:bands}, we compare each band with the reference and compare candidates within that band. The overall task-level C--C results additionally include pairs drawn from different bands. Table~\ref{tab:s3bandcounts} supplies the counts behind the figure, making clear how many comparisons contribute to each rate.

We again see that editing more of the reference does not necessarily produce more misranking. For IOI under mean replacement, R--C failures rise from 2/30 after one edit to 25/30 at 5\% replacement, then fall to 0/30 at 50\%. This pattern is why we examine the full range of edit bands rather than use edit magnitude alone to characterize the failure.
\begin{table}[!tbp]
\centering
\caption{\textbf{Exact KL misranking counts for controlled candidates.} Rows identify tasks; numeric labels are InterpBench IDs. Columns group the validation intervention. R--C denotes reference--candidate pairs; C--C denotes candidate--candidate pairs. Each entry is the number of strict misrankings divided by all eligible pairs, including behavioral and score ties.}\label{tab:s3detail}
\footnotesize
\setlength{\tabcolsep}{3pt}
\renewcommand{\arraystretch}{1.08}
\begin{tabular}{lrrrrrr}
\toprule
\multicolumn{1}{l}{Task} & \multicolumn{2}{c}{Resampling} & \multicolumn{2}{c}{Mean} & \multicolumn{2}{c}{Zero}\\
\cmidrule(lr){2-3}\cmidrule(lr){4-5}\cmidrule(lr){6-7}
 & \multicolumn{1}{r}{R--C} & \multicolumn{1}{r}{C--C} & \multicolumn{1}{r}{R--C} & \multicolumn{1}{r}{C--C} & \multicolumn{1}{r}{R--C} & \multicolumn{1}{r}{C--C}\\
\midrule
\multicolumn{7}{l}{\emph{Human suite}}\\
IOI & 8/180 & 989/16110 & 61/180 & 3973/16110 & 7/180 & 3147/16110\\
Greater-Than & 3/180 & 708/16110 & 3/180 & 997/16110 & 100/180 & 8595/16110\\
Docstring & 13/150 & 625/11175 & 21/150 & 1415/11175 & 44/150 & 2081/11175\\
Acronym & 7/120 & 557/7140 & 19/120 & 1439/7140 & 7/120 & 2411/7140\\
\midrule
\multicolumn{7}{l}{\emph{InterpBench suite}}\\
113 & 4/180 & 247/16110 & 22/180 & 1315/16110 & 21/180 & 1355/16110\\
97 & 0/179 & 718/15931 & 0/179 & 1573/15931 & 0/179 & 2905/15931\\
2 & 3/180 & 107/16110 & 2/180 & 179/16110 & 66/180 & 4040/16110\\
82 & 1/180 & 230/16110 & 1/180 & 626/16110 & 17/180 & 3854/16110\\
111 & 9/180 & 482/16110 & 28/180 & 1234/16110 & 139/180 & 11317/16110\\
45 & 0/180 & 152/16110 & 11/180 & 798/16110 & 21/180 & 1922/16110\\
58 & 5/180 & 376/16110 & 1/180 & 381/16110 & 63/180 & 6435/16110\\
93 & 2/180 & 190/16110 & 0/180 & 301/16110 & 129/180 & 11289/16110\\
103 & 12/180 & 464/16110 & 0/180 & 565/16110 & 27/180 & 3350/16110\\
25 & 14/150 & 1821/11175 & 19/150 & 2410/11175 & 56/150 & 5374/11175\\
\bottomrule
\end{tabular}
\end{table}

\FloatBarrier
\begingroup\scriptsize\setlength{\tabcolsep}{2pt}
\begin{longtable}{llrrrrrr}
\caption{\textbf{KL misranking counts within edit bands.} Band gives one replaced component or the percentage of reference components replaced at fixed size. R--C denotes reference--candidate pairs; C--C denotes candidate--candidate pairs. C--C pairs draw both candidates from the same band; merged bands repeat the same masks. Res. denotes resampling; Mean and Zero denote mean and zero ablation. Each entry is the number of strict misrankings divided by all eligible pairs, including behavioral and score ties. A dash denotes an unavailable pool.}\label{tab:s3bandcounts}\\
\toprule
Task & Band & Res. R--C & Res. C--C & Mean R--C & Mean C--C & Zero R--C & Zero C--C\\
\midrule\endfirsthead
\caption[]{Controlled KL misranking by edit band (continued).}\\
\toprule
Task & Band & Res. R--C & Res. C--C & Mean R--C & Mean C--C & Zero R--C & Zero C--C\\
\midrule\endhead
\bottomrule\endfoot
\bottomrule\endlastfoot
IOI & 1 edit & 5/30 & 112/435 & 2/30 & 75/435 & 2/30 & 51/435\\
IOI & 5\% & 2/30 & 68/435 & 25/30 & 286/435 & 4/30 & 149/435\\
IOI & 10\% & 1/30 & 61/435 & 24/30 & 195/435 & 1/30 & 173/435\\
IOI & 20\% & 0/30 & 65/435 & 10/30 & 64/435 & 0/30 & 192/435\\
IOI & 50\% & 0/30 & 36/435 & 0/30 & 108/435 & 0/30 & 218/435\\
IOI & 75\% & 0/30 & 55/435 & 0/30 & 143/435 & 0/30 & 154/435\\
Greater-Than & 1 edit & 2/30 & 48/435 & 2/30 & 57/435 & 2/30 & 58/435\\
Greater-Than & 5\% & 0/30 & 72/435 & 0/30 & 69/435 & 14/30 & 181/435\\
Greater-Than & 10\% & 1/30 & 87/435 & 1/30 & 92/435 & 20/30 & 170/435\\
Greater-Than & 20\% & 0/30 & 23/435 & 0/30 & 31/435 & 18/30 & 200/435\\
Greater-Than & 50\% & 0/30 & 20/435 & 0/30 & 57/435 & 26/30 & 234/435\\
Greater-Than & 75\% & 0/30 & 22/435 & 0/30 & 18/435 & 20/30 & 44/435\\
Docstring & 1 edit & 11/30 & 74/435 & 10/30 & 120/435 & 23/30 & 244/435\\
Docstring & 5\% & 11/30 & 74/435 & 10/30 & 120/435 & 23/30 & 244/435\\
Docstring & 10\% & 0/30 & 45/435 & 6/30 & 112/435 & 12/30 & 157/435\\
Docstring & 20\% & 2/30 & 27/435 & 5/30 & 86/435 & 8/30 & 105/435\\
Docstring & 50\% & 0/30 & 73/435 & 0/30 & 77/435 & 1/30 & 120/435\\
Docstring & 75\% & 0/30 & 25/435 & 0/30 & 46/435 & 0/30 & 35/435\\
Acronym & 1 edit & 6/30 & 72/435 & 15/30 & 175/435 & 4/30 & 184/435\\
Acronym & 5\% & -- & -- & -- & -- & -- & --\\
Acronym & 10\% & 6/30 & 72/435 & 15/30 & 175/435 & 4/30 & 184/435\\
Acronym & 20\% & 1/30 & 70/435 & 3/30 & 196/435 & 1/30 & 217/435\\
Acronym & 50\% & 0/30 & 36/435 & 1/30 & 128/435 & 2/30 & 274/435\\
Acronym & 75\% & 0/30 & 60/435 & 0/30 & 88/435 & 0/30 & 250/435\\
113 & 1 edit & 0/30 & 0/435 & 0/30 & 0/435 & 0/30 & 0/435\\
113 & 5\% & 3/30 & 43/435 & 9/30 & 131/435 & 8/30 & 131/435\\
113 & 10\% & 1/30 & 11/435 & 11/30 & 104/435 & 11/30 & 95/435\\
113 & 20\% & 0/30 & 11/435 & 2/30 & 41/435 & 2/30 & 53/435\\
113 & 50\% & 0/30 & 3/435 & 0/30 & 9/435 & 0/30 & 14/435\\
113 & 75\% & 0/30 & 0/435 & 0/30 & 0/435 & 0/30 & 0/435\\
97 & 1 edit & 0/29 & 0/406 & 0/29 & 4/406 & 0/29 & 10/406\\
97 & 5\% & 0/30 & 26/435 & 0/30 & 64/435 & 0/30 & 56/435\\
97 & 10\% & 0/30 & 49/435 & 0/30 & 58/435 & 0/30 & 74/435\\
97 & 20\% & 0/30 & 79/435 & 0/30 & 125/435 & 0/30 & 198/435\\
97 & 50\% & 0/30 & 26/435 & 0/30 & 68/435 & 0/30 & 172/435\\
97 & 75\% & 0/30 & 0/435 & 0/30 & 0/435 & 0/30 & 55/435\\
2 & 1 edit & 1/30 & 23/435 & 1/30 & 24/435 & 3/30 & 84/435\\
2 & 5\% & 1/30 & 3/435 & 0/30 & 12/435 & 19/30 & 207/435\\
2 & 10\% & 0/30 & 4/435 & 0/30 & 6/435 & 19/30 & 168/435\\
2 & 20\% & 1/30 & 0/435 & 1/30 & 9/435 & 20/30 & 151/435\\
2 & 50\% & 0/30 & 3/435 & 0/30 & 6/435 & 5/30 & 65/435\\
2 & 75\% & 0/30 & 0/435 & 0/30 & 0/435 & 0/30 & 0/435\\
82 & 1 edit & 0/30 & 1/435 & 0/30 & 1/435 & 1/30 & 26/435\\
82 & 5\% & 1/30 & 23/435 & 0/30 & 70/435 & 9/30 & 191/435\\
82 & 10\% & 0/30 & 14/435 & 1/30 & 50/435 & 2/30 & 111/435\\
82 & 20\% & 0/30 & 23/435 & 0/30 & 53/435 & 5/30 & 203/435\\
82 & 50\% & 0/30 & 7/435 & 0/30 & 14/435 & 0/30 & 120/435\\
82 & 75\% & 0/30 & 0/435 & 0/30 & 0/435 & 0/30 & 1/435\\
111 & 1 edit & 1/30 & 20/435 & 10/30 & 199/435 & 9/30 & 198/435\\
111 & 5\% & 7/30 & 60/435 & 12/30 & 106/435 & 22/30 & 298/435\\
111 & 10\% & 1/30 & 13/435 & 4/30 & 29/435 & 23/30 & 320/435\\
111 & 20\% & 0/30 & 23/435 & 2/30 & 16/435 & 25/30 & 317/435\\
111 & 50\% & 0/30 & 3/435 & 0/30 & 6/435 & 30/30 & 105/435\\
111 & 75\% & 0/30 & 0/435 & 0/30 & 0/435 & 30/30 & 0/435\\
45 & 1 edit & 0/30 & 0/435 & 4/30 & 94/435 & 5/30 & 122/435\\
45 & 5\% & 0/30 & 8/435 & 3/30 & 40/435 & 7/30 & 101/435\\
45 & 10\% & 0/30 & 11/435 & 2/30 & 35/435 & 6/30 & 93/435\\
45 & 20\% & 0/30 & 16/435 & 2/30 & 31/435 & 3/30 & 85/435\\
45 & 50\% & 0/30 & 10/435 & 0/30 & 29/435 & 0/30 & 50/435\\
45 & 75\% & 0/30 & 0/435 & 0/30 & 0/435 & 0/30 & 1/435\\
58 & 1 edit & 1/30 & 21/435 & 0/30 & 0/435 & 3/30 & 73/435\\
58 & 5\% & 3/30 & 35/435 & 0/30 & 7/435 & 8/30 & 123/435\\
58 & 10\% & 1/30 & 11/435 & 1/30 & 19/435 & 9/30 & 136/435\\
58 & 20\% & 0/30 & 15/435 & 0/30 & 39/435 & 11/30 & 132/435\\
58 & 50\% & 0/30 & 25/435 & 0/30 & 32/435 & 13/30 & 177/435\\
58 & 75\% & 0/30 & 0/435 & 0/30 & 0/435 & 19/30 & 36/435\\
93 & 1 edit & 0/30 & 2/435 & 0/30 & 0/435 & 10/30 & 224/435\\
93 & 5\% & 0/30 & 5/435 & 0/30 & 19/435 & 17/30 & 279/435\\
93 & 10\% & 2/30 & 9/435 & 0/30 & 27/435 & 18/30 & 278/435\\
93 & 20\% & 0/30 & 10/435 & 0/30 & 28/435 & 25/30 & 229/435\\
93 & 50\% & 0/30 & 12/435 & 0/30 & 12/435 & 29/30 & 185/435\\
93 & 75\% & 0/30 & 2/435 & 0/30 & 2/435 & 30/30 & 61/435\\
103 & 1 edit & 3/30 & 56/435 & 0/30 & 3/435 & 7/30 & 146/435\\
103 & 5\% & 5/30 & 33/435 & 0/30 & 26/435 & 11/30 & 184/435\\
103 & 10\% & 4/30 & 24/435 & 0/30 & 49/435 & 7/30 & 119/435\\
103 & 20\% & 0/30 & 14/435 & 0/30 & 52/435 & 2/30 & 147/435\\
103 & 50\% & 0/30 & 8/435 & 0/30 & 36/435 & 0/30 & 223/435\\
103 & 75\% & 0/30 & 1/435 & 0/30 & 1/435 & 0/30 & 49/435\\
25 & 1 edit & 6/30 & 135/435 & 4/30 & 97/435 & 5/30 & 127/435\\
25 & 5\% & 4/30 & 77/435 & 9/30 & 131/435 & 10/30 & 175/435\\
25 & 10\% & 3/30 & 86/435 & 5/30 & 103/435 & 6/30 & 166/435\\
25 & 20\% & 1/30 & 80/435 & 1/30 & 170/435 & 11/30 & 260/435\\
25 & 50\% & 0/30 & 51/435 & 0/30 & 142/435 & 24/30 & 135/435\\
25 & 75\% & -- & -- & -- & -- & -- & --\\
\end{longtable}
\endgroup

\clearpage
\begin{table}[!t]
\centering
\caption{\textbf{Misranking across metrics in controlled candidates.} Columns group the validation intervention. R--C denotes reference--candidate pairs; C--C denotes candidate--candidate pairs. We pool all fourteen tasks using only pair identities common to the three metrics. Each entry gives the misranking rate (\%), then the mean behavioral deficit among reversals in parentheses (percentage points, pp). Rate denominators include ties. A dash in parentheses means no reversals; a bare dash means no eligible pairs.}\label{tab:metrics3}
\footnotesize
\setlength{\tabcolsep}{3pt}
\renewcommand{\arraystretch}{1.08}
\begin{tabular}{lrrrrrr}
\toprule
\multicolumn{1}{l}{Metric} & \multicolumn{2}{c}{Resampling} & \multicolumn{2}{c}{Mean} & \multicolumn{2}{c}{Zero}\\
\cmidrule(lr){2-3}\cmidrule(lr){4-5}\cmidrule(lr){6-7}
 & \multicolumn{1}{r}{R--C} & \multicolumn{1}{r}{C--C} & \multicolumn{1}{r}{R--C} & \multicolumn{1}{r}{C--C} & \multicolumn{1}{r}{R--C} & \multicolumn{1}{r}{C--C}\\
\midrule
KL & 3.4 (0.45) & 3.7 (2.91) & 7.8 (3.50) & 8.3 (7.38) & 29.1 (30.05) & 33.0 (28.79)\\
LD & 6.0 (1.18) & 6.1 (2.63) & 7.1 (2.45) & 7.8 (4.76) & 17.3 (13.35) & 18.6 (16.81)\\
PD & 6.5 (11.42) & 5.2 (8.32) & 4.4 (1.09) & 7.8 (7.70) & 25.6 (20.51) & 24.9 (20.57)\\
\bottomrule
\end{tabular}
\end{table}

\FloatBarrier
\begingroup\scriptsize\setlength{\tabcolsep}{2pt}
\begin{longtable}{llrrrrrr}
\caption{\textbf{Logit- and probability-difference misranking in controlled candidates.} Metric is logit difference (LD) or probability difference (PD). R--C denotes reference--candidate pairs; C--C denotes candidate--candidate pairs. Res. denotes resampling; Mean and Zero denote mean and zero ablation. Each entry is the number of strict misrankings divided by all eligible pairs, including behavioral and score ties. Numeric task labels are InterpBench IDs; a dash denotes an unavailable pool.}\label{tab:s3margincounts}\\
\toprule
Task & Metric & Res. R--C & Res. C--C & Mean R--C & Mean C--C & Zero R--C & Zero C--C\\
\midrule\endfirsthead
\caption[]{Controlled LD/PD misranking counts (continued).}\\
\toprule
Task & Metric & Res. R--C & Res. C--C & Mean R--C & Mean C--C & Zero R--C & Zero C--C\\
\midrule\endhead
\bottomrule\endfoot
\bottomrule\endlastfoot
IOI & LD & 4/180 & 467/16110 & 9/180 & 1344/16110 & 142/180 & 12671/16110\\
IOI & PD & 4/180 & 1584/16110 & 7/180 & 1264/16110 & 143/180 & 11993/16110\\
Greater-Than & LD & 11/180 & 1477/16110 & 14/180 & 1788/16110 & 17/180 & 3971/16110\\
Greater-Than & PD & 0/180 & 437/16110 & 0/180 & 759/16110 & 33/180 & 4783/16110\\
Docstring & LD & 12/150 & 706/11175 & 22/150 & 1566/11175 & 32/150 & 1879/11175\\
Docstring & PD & 109/150 & 5906/11175 & 22/150 & 5946/11175 & 119/150 & 8406/11175\\
Acronym & LD & 2/120 & 230/7140 & 5/120 & 388/7140 & 15/120 & 1444/7140\\
Acronym & PD & 1/120 & 313/7140 & 10/120 & 568/7140 & 38/120 & 2265/7140\\
113 & LD & 12/180 & 740/16110 & 13/180 & 809/16110 & 12/180 & 782/16110\\
113 & PD & 7/180 & 370/16110 & 11/180 & 704/16110 & 10/180 & 736/16110\\
97 & LD & 1/179 & 658/15931 & 1/179 & 1330/15931 & 0/179 & 713/15931\\
97 & PD & 0/179 & 206/15931 & 0/179 & 1104/15931 & 0/179 & 883/15931\\
2 & LD & 4/180 & 199/16110 & 5/180 & 243/16110 & 7/180 & 550/16110\\
2 & PD & 4/180 & 134/16110 & 2/180 & 176/16110 & 22/180 & 1557/16110\\
82 & LD & 4/180 & 312/16110 & 6/180 & 486/16110 & 3/180 & 526/16110\\
82 & PD & 1/180 & 63/16110 & 1/180 & 545/16110 & 15/180 & 1609/16110\\
111 & LD & 37/180 & 3861/16110 & 32/180 & 3761/16110 & 130/180 & 9319/16110\\
111 & PD & 4/180 & 180/16110 & 28/180 & 1205/16110 & 102/180 & 8257/16110\\
45 & LD & 3/180 & 318/16110 & 6/180 & 462/16110 & 7/180 & 604/16110\\
45 & PD & 0/180 & 47/16110 & 4/180 & 502/16110 & 20/180 & 1494/16110\\
58 & LD & 5/180 & 397/16110 & 6/180 & 445/16110 & 7/180 & 790/16110\\
58 & PD & 3/180 & 148/16110 & 1/180 & 341/16110 & 25/180 & 2185/16110\\
93 & LD & 10/180 & 415/16110 & 12/180 & 470/16110 & 13/180 & 979/16110\\
93 & PD & 1/180 & 70/16110 & 0/180 & 273/16110 & 15/180 & 1273/16110\\
103 & LD & 9/180 & 433/16110 & 8/180 & 417/16110 & 9/180 & 812/16110\\
103 & PD & 5/180 & 148/16110 & 0/180 & 523/16110 & 17/180 & 1357/16110\\
25 & LD & 31/150 & 2296/11175 & 31/150 & 2529/11175 & 21/150 & 3399/11175\\
25 & PD & 18/150 & 1088/11175 & 20/150 & 2281/11175 & 55/150 & 4664/11175\\
\end{longtable}
\endgroup

\clearpage
\begin{table}[!tbp]
\centering
\caption{\textbf{Behavioral and score gaps among controlled circuit misrankings.} Rows group suite, validation intervention, and pair type (R--C: reference--candidate; C--C: candidate--candidate). Failures counts strict KL reversals. $\Delta Q$ is better minus worse held-out behavior in percentage points (pp); $\Delta F$ is worse minus better validation faithfulness in natural-log units. Both columns show median [25th, 75th percentiles]. Deficit $\geq5$ pp counts failures with at least that behavioral loss. Pairs may share circuits.}\label{tab:s3gapdist}
\scriptsize
\setlength{\tabcolsep}{3pt}
\begin{tabular}{lllrrrr}
\toprule
Suite & Intervention & Pair & Failures & $\Delta Q$ (pp) & $\Delta F$ & Deficit $\geq5$ pp\\
\midrule
\emph{Human suite} & Resampling & R--C & 31 & 0.33 [0.33, 1.33] & 0.017 [0.013, 0.054] & 0\\
\emph{Human suite} & Resampling & C--C & 2879 & 1.00 [0.33, 2.33] & 0.070 [0.027, 0.157] & 333\\
\emph{Human suite} & Mean & R--C & 104 & 2.33 [1.33, 7.08] & 0.226 [0.034, 0.421] & 35\\
\emph{Human suite} & Mean & C--C & 7824 & 3.00 [1.00, 8.67] & 0.161 [0.049, 0.359] & 2968\\
\emph{Human suite} & Zero & R--C & 158 & 6.33 [1.33, 45.17] & 0.883 [0.137, 2.629] & 86\\
\emph{Human suite} & Zero & C--C & 16234 & 14.11 [2.33, 42.33] & 1.162 [0.315, 2.885] & 10607\\
\emph{InterpBench suite} & Resampling & R--C & 50 & 0.11 [0.07, 0.33] & 0.005 [0.000, 0.014] & 0\\
\emph{InterpBench suite} & Resampling & C--C & 4787 & 0.52 [0.11, 4.56] & 0.243 [0.010, 2.976] & 1138\\
\emph{InterpBench suite} & Mean & R--C & 84 & 0.41 [0.18, 0.87] & 0.000 [0.000, 0.001] & 1\\
\emph{InterpBench suite} & Mean & C--C & 9382 & 1.54 [0.33, 9.93] & 0.036 [0.000, 1.202] & 3399\\
\emph{InterpBench suite} & Zero & R--C & 539 & 7.15 [1.94, 78.19] & 2.030 [0.319, 48.346] & 364\\
\emph{InterpBench suite} & Zero & C--C & 51841 & 13.67 [3.52, 60.48] & 3.530 [0.708, 30.387] & 36211\\
\bottomrule
\end{tabular}
\end{table}

\FloatBarrier
We report LD and PD counts for each task's candidate pool in Table~\ref{tab:s3margincounts}, and compare all three metrics on their common pair identities in Table~\ref{tab:metrics3}. We also show the distribution of KL behavioral deficits in Table~\ref{tab:s3gapdist}. For Human C--C pairs under zero replacement, we find both near-tied reversals and failures with substantial behavioral losses. The distribution makes this variation visible in a way that a rate or mean alone cannot. We examine which failures the metrics share in Appendix~\ref{app:overlap}.

\section{Section 4: discovery setup, executable-size control, and additional results}
\label{app:section4}
\subsection{Discovery objectives and budgets}
We use the released implementations of EAP, EAP-IG, ACDC, and Edge-SP \citep{syed,hanna,conmy}, with the task adaptations and size-controlled extraction described below. For each task and seed, we give all four methods the same 100 discovery prompts. We use seeds 0--19 and 810--819 and keep these prompts disjoint from evaluation. We then evaluate the returned circuits on the Section 3 validation and held-out prompts, using the same mean bank, scored positions, and three donor draws. This lets us compare discovered candidates under the same conditions as the controlled candidates.

For Human EAP-IG, we use 30 integration steps. For InterpBench attribution, we use categorical KL as the discovery objective, batches of ten prompts, and 30 EAP-IG steps. We keep discovery fixed when we later evaluate the candidates with LD or PD; those comparisons change the score, not the circuits returned by the method.

We cap each discovery instance at 30 minutes. ACDC follows its sequential deletion trajectory, from which we extract a reference-size circuit using the recorded pruning order. For Edge-SP, we freeze the model weights and optimize mask parameters with Adam at learning rate 0.01 and sparsity coefficient 1. We allow at most 300 updates over the full set of 100 prompts, using microbatches of four and sharing the sampled mask across the microbatches of each update. All 120 Edge-SP task--seed instances for InterpBench 113, IOI, Greater-Than, and Acronym complete 300 updates before the time limit. We use FP32 for native mask optimization and FP64 for InterpBench evaluation. For Acronym, we adapt the masks to attention-head outputs to preserve the task's head-level circuit representation.

\subsection{Candidate extraction and actual-size eligibility}
We extract each candidate at its validated reference size, using the component universe described in Appendix~\ref{app:section3}. For EAP, we order full-universe scores by absolute attribution, breaking ties with canonical edge names; for EAP-IG, we use the corresponding integrated scores. For ACDC, we replay accepted native deletions and use the first checkpoint at the required size. For Edge-SP, we use the discovery-objective checkpoint and its top-$K$ retention parameters. We do not use validation scores or Q to choose extraction checkpoints.

Before evaluation, we check each mask's executable keys, actual component count, identity, and finite outputs against the validated representation. All 1,680 returned circuits have the exact reference size. We then collapse duplicate masks before forming pairs. Repeated discovery of the same circuit still counts as a completed run, but it does not create an additional distinct candidate or pair.

Our Human ACDC pool includes IOI and Greater-Than as well as Docstring and Acronym. Even when an instance reaches the discovery budget, the extraction procedure yields an exact-size circuit. In the result tables, we use a dash when deduplication leaves no available pair pool, and zero when pairs are available but none misrank.

\subsection{Reference and discovered-candidate performance}
We report candidate Q in Table~\ref{tab:s4perf}, averaging over the unique masks for each task and method. Every candidate has the task's reference size, and a mask receives the same weight regardless of how often it appears in pairwise comparisons. For Greater-Than, we retain semantic Q throughout, including when evaluating ACDC and Edge-SP outputs with LD or PD. We match each margin score to the Q value of that same circuit before determining its pairwise ordering.

We observe substantial variation in candidate behavior across tasks and methods. These means tell us what circuits are available to choose from. The misranking analysis asks the next question: given those choices, does faithfulness prefer the better one?
\FloatBarrier

\subsection{Complete task--method misranking and deficit summaries}
We report KL and LD/PD counts by task, method, intervention, and pair type in Tables~\ref{tab:s4detail} and \ref{tab:s4margincounts}. We retain Q and score ties in the total-pair denominators. Method-level summaries describe each method's own candidates, while cross-metric summaries use only the exact pair identities shared by the metrics.

We also examine the distribution of behavioral deficits in Table~\ref{tab:s4gapdist}. Frequent reversals do not always mean large losses, and a lower rate can still include costly choices. Reporting both helps distinguish a score that often switches near-tied candidates from one that sometimes prefers a substantially worse circuit.
\FloatBarrier
\begingroup\scriptsize\setlength{\tabcolsep}{2pt}
\begin{longtable}{llrrrrrr}
\caption{\textbf{Exact KL misranking counts for discovery outputs.} Task identifies the behavior or InterpBench ID; Method identifies the discovery algorithm. R--C denotes reference--candidate pairs; C--C denotes candidate--candidate pairs. Res. denotes resampling; Mean and Zero denote mean and zero ablation. Each entry is the number of strict misrankings divided by all eligible pairs, including behavioral and score ties. All pairs match the reference\textquotesingle s executable size; a dash denotes an unavailable pool after deduplication.}\label{tab:s4detail}\\
\toprule
Task & Method & Res. R--C & Res. C--C & Mean R--C & Mean C--C & Zero R--C & Zero C--C\\
\midrule\endfirsthead
\caption[]{Discovery-output KL misranking counts (continued).}\\
\toprule
Task & Method & Res. R--C & Res. C--C & Mean R--C & Mean C--C & Zero R--C & Zero C--C\\
\midrule\endhead
\bottomrule\endfoot
\bottomrule\endlastfoot
IOI & EAP & 0/30 & 119/435 & 29/30 & 190/435 & 0/30 & 218/435\\
IOI & EAP-IG & 23/30 & 121/435 & 23/30 & 271/435 & 4/30 & 208/435\\
IOI & ACDC & 0/30 & 106/435 & 30/30 & 98/435 & 0/30 & 186/435\\
IOI & Edge-SP & 0/30 & 41/435 & 14/30 & 64/435 & 1/30 & 185/435\\
Greater-Than & EAP & 1/30 & 23/435 & 1/30 & 30/435 & 21/30 & 200/435\\
Greater-Than & EAP-IG & 7/30 & 100/435 & 17/30 & 168/435 & 12/30 & 150/435\\
Greater-Than & ACDC & 0/30 & 49/435 & 0/30 & 63/435 & 30/30 & 86/435\\
Greater-Than & Edge-SP & 17/30 & 73/435 & 20/30 & 80/435 & 5/30 & 162/435\\
Docstring & EAP & 3/24 & 41/276 & 1/24 & 45/276 & 4/24 & 91/276\\
Docstring & EAP-IG & 9/28 & 85/378 & 12/28 & 137/378 & 10/28 & 109/378\\
Docstring & ACDC & 4/24 & 79/276 & 11/24 & 125/276 & 17/24 & 187/276\\
Docstring & Edge-SP & 1/30 & 89/435 & 21/30 & 116/435 & 12/30 & 173/435\\
Acronym & EAP & 1/6 & 2/15 & 4/6 & 3/15 & 0/6 & 3/15\\
Acronym & EAP-IG & 0/4 & 0/6 & 0/4 & 3/6 & 0/4 & 4/6\\
Acronym & ACDC & 6/6 & 5/15 & 6/6 & 7/15 & 0/6 & 4/15\\
Acronym & Edge-SP & 0/30 & 26/435 & 1/30 & 67/435 & 1/30 & 226/435\\
113 & EAP & 0/30 & 9/435 & 3/30 & 26/435 & 4/30 & 138/435\\
113 & EAP-IG & 0/30 & 0/435 & 0/30 & 0/435 & 0/30 & 0/435\\
113 & ACDC & 0/30 & 45/435 & 0/30 & 107/435 & 0/30 & 104/435\\
113 & Edge-SP & 0/30 & 88/435 & 0/30 & 181/435 & 0/30 & 163/435\\
97 & EAP & 0/30 & 129/435 & 0/30 & 130/435 & 0/30 & 169/435\\
97 & EAP-IG & 0/30 & 0/435 & 0/30 & 0/435 & 0/30 & 0/435\\
97 & ACDC & 0/5 & 0/10 & 0/5 & 2/10 & 0/5 & 3/10\\
97 & Edge-SP & 0/30 & 0/435 & 1/30 & 0/435 & 0/30 & 0/435\\
2 & EAP & 0/30 & 0/435 & 0/30 & 0/435 & 29/30 & 66/435\\
2 & EAP-IG & 0/30 & 0/435 & 0/30 & 0/435 & 30/30 & 0/435\\
2 & ACDC & 0/16 & 3/120 & 14/16 & 4/120 & 14/16 & 41/120\\
2 & Edge-SP & 0/30 & 146/435 & 0/30 & 0/435 & 30/30 & 118/435\\
82 & EAP & 0/30 & 7/435 & 1/30 & 3/435 & 3/30 & 110/435\\
82 & EAP-IG & 0/30 & 0/435 & 0/30 & 0/435 & 1/30 & 0/435\\
82 & ACDC & 0/30 & 31/435 & 0/30 & 203/435 & 30/30 & 350/435\\
82 & Edge-SP & 0/30 & 70/435 & 0/30 & 152/435 & 0/30 & 268/435\\
111 & EAP & 10/30 & 77/435 & 8/30 & 109/435 & 17/30 & 231/435\\
111 & EAP-IG & 0/30 & 90/435 & 1/30 & 60/435 & 28/30 & 170/435\\
111 & ACDC & 6/30 & 65/435 & 9/30 & 116/435 & 9/30 & 271/435\\
111 & Edge-SP & 0/30 & 78/435 & 0/30 & 61/435 & 30/30 & 303/435\\
45 & EAP & 1/30 & 30/435 & 10/30 & 88/435 & 10/30 & 164/435\\
45 & EAP-IG & 0/30 & 17/435 & 19/30 & 38/435 & 0/30 & 48/435\\
45 & ACDC & 4/29 & 24/406 & 3/29 & 74/406 & 15/29 & 205/406\\
45 & Edge-SP & 0/30 & 74/435 & 0/30 & 114/435 & 0/30 & 203/435\\
58 & EAP & 1/30 & 30/435 & 1/30 & 3/435 & 24/30 & 161/435\\
58 & EAP-IG & 0/25 & 0/300 & 0/25 & 0/300 & 25/25 & 0/300\\
58 & ACDC & 1/15 & 6/105 & 5/15 & 29/105 & 6/15 & 44/105\\
58 & Edge-SP & 1/30 & 2/435 & 0/30 & 0/435 & 30/30 & 38/435\\
93 & EAP & 0/30 & 1/435 & 1/30 & 1/435 & 16/30 & 65/435\\
93 & EAP-IG & 0/30 & 0/435 & 0/30 & 0/435 & 18/30 & 0/435\\
93 & ACDC & 0/12 & 1/66 & 0/12 & 9/66 & 0/12 & 22/66\\
93 & Edge-SP & 0/30 & 36/435 & 0/30 & 0/435 & 0/30 & 188/435\\
103 & EAP & 2/30 & 9/435 & 2/30 & 10/435 & 25/30 & 99/435\\
103 & EAP-IG & 0/30 & 0/435 & 0/30 & 0/435 & 30/30 & 0/435\\
103 & ACDC & 7/22 & 7/231 & 0/22 & 23/231 & 2/22 & 119/231\\
103 & Edge-SP & 0/30 & 113/435 & 0/30 & 0/435 & 30/30 & 198/435\\
25 & EAP & 8/27 & 15/351 & 9/27 & 91/351 & 14/27 & 55/351\\
25 & EAP-IG & 1/30 & 0/435 & 4/30 & 4/435 & 0/30 & 22/435\\
25 & ACDC & 3/6 & 8/15 & 4/6 & 12/15 & 2/6 & 12/15\\
25 & Edge-SP & 5/30 & 99/435 & 11/30 & 163/435 & 7/30 & 214/435\\
\end{longtable}
\endgroup

\FloatBarrier
\begingroup\scriptsize\setlength{\tabcolsep}{2pt}
\begin{longtable}{lllrrrrrr}
\caption{\textbf{Logit- and probability-difference misranking in discovery outputs.} Task identifies the behavior or InterpBench ID; Method identifies the discovery algorithm; Metric is logit difference (LD) or probability difference (PD). R--C denotes reference--candidate pairs; C--C denotes candidate--candidate pairs. Res. denotes resampling; Mean and Zero denote mean and zero ablation. Each entry is the number of strict misrankings divided by all eligible pairs, including behavioral and score ties. A dash denotes an unavailable pair pool.}\label{tab:s4margincounts}\\
\toprule
Task & Method & Metric & Res. R--C & Res. C--C & Mean R--C & Mean C--C & Zero R--C & Zero C--C\\
\midrule\endfirsthead
\caption[]{Discovery-output LD/PD misranking counts (continued).}\\
\toprule
Task & Method & Metric & Res. R--C & Res. C--C & Mean R--C & Mean C--C & Zero R--C & Zero C--C\\
\midrule\endhead
\bottomrule\endfoot
\bottomrule\endlastfoot
IOI & EAP & LD & 0/30 & 83/435 & 0/30 & 221/435 & 30/30 & 212/435\\
IOI & EAP & PD & 0/30 & 115/435 & 0/30 & 202/435 & 30/30 & 193/435\\
IOI & EAP-IG & LD & 4/30 & 98/435 & 4/30 & 134/435 & 23/30 & 149/435\\
IOI & EAP-IG & PD & 4/30 & 109/435 & 4/30 & 119/435 & 23/30 & 154/435\\
IOI & ACDC & LD & 0/30 & 77/435 & 0/30 & 96/435 & 30/30 & 132/435\\
IOI & ACDC & PD & 0/30 & 102/435 & 0/30 & 111/435 & 30/30 & 137/435\\
IOI & Edge-SP & LD & 0/30 & 21/435 & 0/30 & 26/435 & 30/30 & 293/435\\
IOI & Edge-SP & PD & 0/30 & 34/435 & 0/30 & 47/435 & 30/30 & 184/435\\
Greater-Than & EAP & LD & 15/30 & 98/435 & 14/30 & 162/435 & 1/30 & 177/435\\
Greater-Than & EAP & PD & 1/30 & 31/435 & 1/30 & 89/435 & 1/30 & 206/435\\
Greater-Than & EAP-IG & LD & 7/30 & 88/435 & 16/30 & 148/435 & 17/30 & 130/435\\
Greater-Than & EAP-IG & PD & 6/30 & 66/435 & 17/30 & 151/435 & 17/30 & 124/435\\
Greater-Than & ACDC & LD & 30/30 & 53/435 & 30/30 & 87/435 & 0/30 & 136/435\\
Greater-Than & ACDC & PD & 0/30 & 41/435 & 0/30 & 62/435 & 0/30 & 112/435\\
Greater-Than & Edge-SP & LD & 5/30 & 50/435 & 15/30 & 82/435 & 20/30 & 121/435\\
Greater-Than & Edge-SP & PD & 5/30 & 18/435 & 20/30 & 37/435 & 20/30 & 92/435\\
Docstring & EAP & LD & 13/24 & 48/276 & 1/24 & 68/276 & 5/24 & 63/276\\
Docstring & EAP & PD & 23/24 & 106/276 & 1/24 & 169/276 & 16/24 & 150/276\\
Docstring & EAP-IG & LD & 8/28 & 108/378 & 18/28 & 143/378 & 9/28 & 174/378\\
Docstring & EAP-IG & PD & 15/28 & 123/378 & 18/28 & 155/378 & 17/28 & 222/378\\
Docstring & ACDC & LD & 3/24 & 65/276 & 11/24 & 117/276 & 16/24 & 149/276\\
Docstring & ACDC & PD & 3/24 & 90/276 & 11/24 & 118/276 & 6/24 & 90/276\\
Docstring & Edge-SP & LD & 2/30 & 50/435 & 21/30 & 102/435 & 21/30 & 178/435\\
Docstring & Edge-SP & PD & 8/30 & 74/435 & 21/30 & 311/435 & 17/30 & 235/435\\
Acronym & EAP & LD & 0/6 & 1/15 & 0/6 & 5/15 & 0/6 & 5/15\\
Acronym & EAP & PD & 0/6 & 2/15 & 0/6 & 5/15 & 6/6 & 9/15\\
Acronym & EAP-IG & LD & 0/4 & 1/6 & 0/4 & 3/6 & 0/4 & 3/6\\
Acronym & EAP-IG & PD & 0/4 & 1/6 & 0/4 & 3/6 & 3/4 & 2/6\\
Acronym & ACDC & LD & 0/6 & 3/15 & 0/6 & 7/15 & 0/6 & 1/15\\
Acronym & ACDC & PD & 4/6 & 2/15 & 0/6 & 4/15 & 1/6 & 13/15\\
Acronym & Edge-SP & LD & 0/30 & 26/435 & 0/30 & 34/435 & 0/30 & 106/435\\
Acronym & Edge-SP & PD & 0/30 & 23/435 & 0/30 & 32/435 & 0/30 & 199/435\\
113 & EAP & LD & 0/30 & 17/435 & 0/30 & 127/435 & 0/30 & 165/435\\
113 & EAP & PD & 0/30 & 4/435 & 0/30 & 35/435 & 0/30 & 166/435\\
113 & EAP-IG & LD & 30/30 & 0/435 & 30/30 & 0/435 & 30/30 & 0/435\\
113 & EAP-IG & PD & 0/30 & 0/435 & 30/30 & 0/435 & 30/30 & 0/435\\
113 & ACDC & LD & 0/30 & 87/435 & 0/30 & 100/435 & 0/30 & 101/435\\
113 & ACDC & PD & 0/30 & 33/435 & 0/30 & 103/435 & 0/30 & 103/435\\
113 & Edge-SP & LD & 30/30 & 221/435 & 30/30 & 269/435 & 30/30 & 267/435\\
113 & Edge-SP & PD & 0/30 & 72/435 & 30/30 & 0/435 & 30/30 & 0/435\\
97 & EAP & LD & 0/30 & 213/435 & 0/30 & 202/435 & 0/30 & 160/435\\
97 & EAP & PD & 0/30 & 104/435 & 0/30 & 113/435 & 0/30 & 156/435\\
97 & EAP-IG & LD & 0/30 & 0/435 & 0/30 & 0/435 & 0/30 & 0/435\\
97 & EAP-IG & PD & 0/30 & 0/435 & 0/30 & 0/435 & 0/30 & 0/435\\
97 & ACDC & LD & 0/5 & 2/10 & 0/5 & 2/10 & 0/5 & 2/10\\
97 & ACDC & PD & 0/5 & 0/10 & 0/5 & 1/10 & 0/5 & 1/10\\
97 & Edge-SP & LD & 1/30 & 10/435 & 1/30 & 17/435 & 0/30 & 0/435\\
97 & Edge-SP & PD & 0/30 & 0/435 & 0/30 & 1/435 & 0/30 & 0/435\\
2 & EAP & LD & 0/30 & 0/435 & 0/30 & 0/435 & 29/30 & 42/435\\
2 & EAP & PD & 0/30 & 0/435 & 0/30 & 0/435 & 29/30 & 48/435\\
2 & EAP-IG & LD & 0/30 & 0/435 & 0/30 & 0/435 & 30/30 & 0/435\\
2 & EAP-IG & PD & 0/30 & 0/435 & 0/30 & 0/435 & 30/30 & 0/435\\
2 & ACDC & LD & 2/16 & 12/120 & 2/16 & 12/120 & 14/16 & 12/120\\
2 & ACDC & PD & 3/16 & 1/120 & 14/16 & 4/120 & 14/16 & 27/120\\
2 & Edge-SP & LD & 0/30 & 261/435 & 0/30 & 258/435 & 30/30 & 214/435\\
2 & Edge-SP & PD & 0/30 & 126/435 & 0/30 & 0/435 & 30/30 & 123/435\\
82 & EAP & LD & 3/30 & 17/435 & 6/30 & 22/435 & 2/30 & 45/435\\
82 & EAP & PD & 0/30 & 7/435 & 1/30 & 3/435 & 3/30 & 109/435\\
82 & EAP-IG & LD & 0/30 & 0/435 & 0/30 & 0/435 & 0/30 & 0/435\\
82 & EAP-IG & PD & 0/30 & 0/435 & 0/30 & 0/435 & 0/30 & 0/435\\
82 & ACDC & LD & 18/30 & 79/435 & 17/30 & 86/435 & 18/30 & 76/435\\
82 & ACDC & PD & 0/30 & 16/435 & 0/30 & 203/435 & 30/30 & 306/435\\
82 & Edge-SP & LD & 30/30 & 140/435 & 30/30 & 241/435 & 0/30 & 209/435\\
82 & Edge-SP & PD & 0/30 & 42/435 & 0/30 & 149/435 & 0/30 & 270/435\\
111 & EAP & LD & 14/30 & 151/435 & 16/30 & 113/435 & 16/30 & 150/435\\
111 & EAP & PD & 13/30 & 22/435 & 8/30 & 109/435 & 16/30 & 181/435\\
111 & EAP-IG & LD & 14/30 & 168/435 & 0/30 & 139/435 & 0/30 & 135/435\\
111 & EAP-IG & PD & 0/30 & 65/435 & 1/30 & 60/435 & 26/30 & 170/435\\
111 & ACDC & LD & 7/30 & 213/435 & 8/30 & 171/435 & 9/30 & 273/435\\
111 & ACDC & PD & 2/30 & 42/435 & 9/30 & 106/435 & 9/30 & 265/435\\
111 & Edge-SP & LD & 0/30 & 195/435 & 0/30 & 259/435 & 0/30 & 333/435\\
111 & Edge-SP & PD & 0/30 & 31/435 & 0/30 & 61/435 & 30/30 & 275/435\\
45 & EAP & LD & 6/30 & 68/435 & 7/30 & 86/435 & 8/30 & 130/435\\
45 & EAP & PD & 2/30 & 24/435 & 10/30 & 92/435 & 8/30 & 162/435\\
45 & EAP-IG & LD & 0/30 & 273/435 & 0/30 & 290/435 & 0/30 & 226/435\\
45 & EAP-IG & PD & 0/30 & 21/435 & 19/30 & 35/435 & 0/30 & 42/435\\
45 & ACDC & LD & 9/29 & 55/406 & 12/29 & 55/406 & 14/29 & 109/406\\
45 & ACDC & PD & 0/29 & 23/406 & 7/29 & 57/406 & 19/29 & 227/406\\
45 & Edge-SP & LD & 30/30 & 134/435 & 30/30 & 172/435 & 30/30 & 221/435\\
45 & Edge-SP & PD & 0/30 & 82/435 & 0/30 & 123/435 & 0/30 & 197/435\\
58 & EAP & LD & 1/30 & 50/435 & 2/30 & 48/435 & 24/30 & 133/435\\
58 & EAP & PD & 0/30 & 27/435 & 1/30 & 3/435 & 24/30 & 153/435\\
58 & EAP-IG & LD & 0/25 & 0/300 & 0/25 & 0/300 & 25/25 & 0/300\\
58 & EAP-IG & PD & 0/25 & 0/300 & 0/25 & 0/300 & 25/25 & 0/300\\
58 & ACDC & LD & 5/15 & 10/105 & 5/15 & 9/105 & 6/15 & 28/105\\
58 & ACDC & PD & 2/15 & 3/105 & 5/15 & 29/105 & 6/15 & 38/105\\
58 & Edge-SP & LD & 0/30 & 7/435 & 0/30 & 26/435 & 30/30 & 35/435\\
58 & Edge-SP & PD & 0/30 & 2/435 & 0/30 & 0/435 & 30/30 & 37/435\\
93 & EAP & LD & 1/30 & 32/435 & 1/30 & 32/435 & 26/30 & 6/435\\
93 & EAP & PD & 0/30 & 0/435 & 1/30 & 2/435 & 26/30 & 46/435\\
93 & EAP-IG & LD & 0/30 & 0/435 & 0/30 & 0/435 & 30/30 & 0/435\\
93 & EAP-IG & PD & 0/30 & 0/435 & 0/30 & 0/435 & 30/30 & 0/435\\
93 & ACDC & LD & 5/12 & 11/66 & 5/12 & 11/66 & 0/12 & 25/66\\
93 & ACDC & PD & 0/12 & 3/66 & 0/12 & 9/66 & 0/12 & 30/66\\
93 & Edge-SP & LD & 30/30 & 263/435 & 30/30 & 260/435 & 0/30 & 229/435\\
93 & Edge-SP & PD & 0/30 & 47/435 & 0/30 & 0/435 & 0/30 & 186/435\\
103 & EAP & LD & 0/30 & 30/435 & 0/30 & 36/435 & 25/30 & 115/435\\
103 & EAP & PD & 0/30 & 9/435 & 2/30 & 10/435 & 25/30 & 114/435\\
103 & EAP-IG & LD & 0/30 & 0/435 & 0/30 & 0/435 & 30/30 & 0/435\\
103 & EAP-IG & PD & 0/30 & 0/435 & 0/30 & 0/435 & 30/30 & 0/435\\
103 & ACDC & LD & 2/22 & 25/231 & 1/22 & 25/231 & 2/22 & 44/231\\
103 & ACDC & PD & 1/22 & 11/231 & 0/22 & 23/231 & 2/22 & 115/231\\
103 & Edge-SP & LD & 0/30 & 224/435 & 0/30 & 221/435 & 30/30 & 199/435\\
103 & Edge-SP & PD & 0/30 & 103/435 & 0/30 & 0/435 & 30/30 & 217/435\\
25 & EAP & LD & 13/27 & 83/351 & 6/27 & 66/351 & 14/27 & 88/351\\
25 & EAP & PD & 8/27 & 90/351 & 9/27 & 91/351 & 14/27 & 55/351\\
25 & EAP-IG & LD & 30/30 & 81/435 & 27/30 & 54/435 & 0/30 & 45/435\\
25 & EAP-IG & PD & 1/30 & 1/435 & 4/30 & 4/435 & 0/30 & 22/435\\
25 & ACDC & LD & 0/6 & 5/15 & 0/6 & 3/15 & 6/6 & 11/15\\
25 & ACDC & PD & 2/6 & 8/15 & 4/6 & 12/15 & 2/6 & 12/15\\
25 & Edge-SP & LD & 17/30 & 133/435 & 7/30 & 61/435 & 8/30 & 250/435\\
25 & Edge-SP & PD & 3/30 & 96/435 & 12/30 & 143/435 & 6/30 & 205/435\\
\end{longtable}
\endgroup

\begin{table}[!tbp]
\centering
\caption{\textbf{Behavioral and score gaps among discovered circuit misrankings.} Rows group suite, validation intervention, and pair type (R--C: reference--candidate; C--C: candidate--candidate). Failures counts strict KL reversals. $\Delta Q$ is better minus worse held-out behavior in percentage points (pp); $\Delta F$ is worse minus better validation faithfulness in natural-log units. Both columns show median [25th, 75th percentiles]. Deficit $\geq5$ pp counts failures with at least that behavioral loss. Pairs may share circuits.}\label{tab:s4gapdist}
\scriptsize
\setlength{\tabcolsep}{3pt}
\begin{tabular}{lllrrrr}
\toprule
Suite & Intervention & Pair & Failures & $\Delta Q$ (pp) & $\Delta F$ & Deficit $\geq5$ pp\\
\midrule
\emph{Human suite} & Resampling & R--C & 72 & 0.67 [0.33, 1.83] & 0.076 [0.030, 0.117] & 2\\
\emph{Human suite} & Resampling & C--C & 959 & 0.67 [0.33, 1.67] & 0.040 [0.014, 0.107] & 36\\
\emph{Human suite} & Mean & R--C & 190 & 3.67 [0.67, 5.67] & 0.326 [0.091, 0.485] & 62\\
\emph{Human suite} & Mean & C--C & 1467 & 1.00 [0.67, 2.67] & 0.053 [0.018, 0.127] & 136\\
\emph{Human suite} & Zero & R--C & 117 & 1.00 [0.67, 3.33] & 1.663 [0.388, 3.260] & 18\\
\emph{Human suite} & Zero & C--C & 2192 & 1.67 [0.67, 4.33] & 0.813 [0.300, 1.822] & 506\\
\emph{InterpBench suite} & Resampling & R--C & 50 & 0.33 [0.23, 0.66] & 0.019 [0.009, 0.051] & 0\\
\emph{InterpBench suite} & Resampling & C--C & 1310 & 0.11 [0.04, 0.19] & 0.005 [0.001, 0.017] & 2\\
\emph{InterpBench suite} & Mean & R--C & 106 & 0.63 [0.33, 1.18] & 0.000 [0.000, 0.001] & 0\\
\emph{InterpBench suite} & Mean & C--C & 1813 & 0.33 [0.11, 0.89] & 0.000 [0.000, 0.008] & 24\\
\emph{InterpBench suite} & Zero & R--C & 479 & 1.22 [0.59, 2.22] & 5.035 [0.902, 13.006] & 10\\
\emph{InterpBench suite} & Zero & C--C & 4362 & 0.37 [0.11, 1.19] & 0.380 [0.044, 3.452] & 299\\
\bottomrule
\end{tabular}
\end{table}

\FloatBarrier
We treat these pooled counts as descriptions of the discovered candidate pools because many pairs share a circuit. We report which failures persist across metrics on the same pairs in Appendix~\ref{app:overlap}.

\subsection{Faithfulness metrics and comparison between suites}
\label{app:discoverymetrics}
In Table~\ref{tab:metrics4}, we compare KL, LD, and PD on the same available pairs within each suite. We hold Q, masks, prompts, donors, and intervention fixed, changing only the faithfulness score. As elsewhere, we use semantic Q for Greater-Than and retain ties in the denominator. Each cell gives the fraction of common pairs that misrank and the mean behavioral deficit among those reversals.

We find misranking under all three metrics. For KL, the Human suite has higher pooled C--C rates than InterpBench under resampling, mean, and zero. We do not attribute this contrast to model scale or semi-synthetic construction alone, since the suites also differ in tasks, candidate pools, and tie frequencies. What the comparison does show is that the InterpBench results do not capture the full extent of the failures we observe in pretrained-model circuits. This is why we evaluate both settings.
\FloatBarrier

\begin{table}[!tbp]
\centering
\caption{\textbf{Misranking across metrics in discovery outputs.} Columns group the validation intervention. R--C denotes reference--candidate pairs; C--C denotes candidate--candidate pairs. Each entry gives the misranking rate (\%) and mean behavioral deficit among reversals in parentheses (percentage points, pp). Within each suite, we use common pair identities and hold masks, held-out behavior, prompts, and intervention fixed across metrics. Rate denominators include ties. A dash in parentheses means no reversals; a bare dash means no eligible pairs.}\label{tab:metrics4}
\footnotesize
\setlength{\tabcolsep}{3pt}
\renewcommand{\arraystretch}{1.08}
\begin{tabular}{lrrrrrr}
\toprule
\multicolumn{1}{l}{Metric} & \multicolumn{2}{c}{Resampling} & \multicolumn{2}{c}{Mean} & \multicolumn{2}{c}{Zero}\\
\cmidrule(lr){2-3}\cmidrule(lr){4-5}\cmidrule(lr){6-7}
 & \multicolumn{1}{r}{R--C} & \multicolumn{1}{r}{C--C} & \multicolumn{1}{r}{R--C} & \multicolumn{1}{r}{C--C} & \multicolumn{1}{r}{R--C} & \multicolumn{1}{r}{C--C}\\
\midrule
\multicolumn{7}{l}{\emph{Human suite}}\\
KL & 18.4 (1.27) & 18.0 (1.39) & 48.5 (4.15) & 27.6 (2.28) & 29.8 (3.32) & 41.2 (5.55)\\
LD & 22.2 (1.67) & 16.4 (1.25) & 33.2 (2.30) & 27.0 (1.99) & 51.5 (5.23) & 38.2 (4.60)\\
PD & 17.6 (3.71) & 17.6 (2.01) & 23.7 (2.58) & 30.4 (3.79) & 55.4 (5.33) & 39.9 (5.75)\\
\midrule
\multicolumn{7}{l}{\emph{InterpBench suite}}\\
KL & 4.6 (0.48) & 8.7 (0.20) & 9.8 (0.76) & 12.0 (0.74) & 44.1 (1.51) & 28.9 (2.27)\\
LD & 27.4 (1.87) & 21.7 (0.46) & 25.1 (1.88) & 23.0 (1.42) & 47.5 (1.31) & 27.0 (1.99)\\
PD & 3.4 (0.39) & 7.4 (0.20) & 15.4 (0.94) & 10.5 (0.84) & 51.0 (1.40) & 26.9 (2.41)\\
\bottomrule
\end{tabular}
\end{table}

\FloatBarrier

\subsection{Connection to the context-restoration study}
We draw the 100 persistent misrankings studied in Section~\ref{sec:repair} from these discovery outputs. For each selected case, we keep the candidate circuits, original intervention, evaluation prompts, donor assignments, and held-out Q ordering fixed. We then change only selected context signals during restoration. Appendix~\ref{app:repair} describes the selection and reports every outcome.

\section{Section 5: cohort selection, restoration search, and complete outcomes}
\label{app:repair}
\subsection{Selecting 100 restoration cases}
We define a case by its task, unordered circuit pair, and original intervention. We select cases from the fully evaluated discovery pool in Section~\ref{sec:discovery}, with each instance subject to the 30-minute limit. We require both circuits to have the task's exact reference size, a Q deficit of at least 0.5 percentage points, and a strict KL misranking on both validation and held-out prompts. We exclude Q and score ties and retain the semantic Q criterion for Greater-Than. We also require 100 distinct circuit pairs, so we never count the same pair twice under different interventions.

We first maximize coverage of task--method combinations with eligible persistent failures. This includes all 51 eligible combinations; the other five of the possible 56 have no eligible pair. We then allocate 50 cases to each suite and 50, 25, and 25 cases to resampling, mean, and zero, respectively. Within those constraints, we balance pair types and task representation. The resulting cohort covers all 14 tasks and contains 47 R--C and 53 C--C cases. We fix it before searching for restorations. We give its task allocation in Table~\ref{tab:s5allocation} and list all cases in Table~\ref{tab:s5catalog}.
\begin{table}[t]
\centering
\caption{\textbf{Restoration cases by task.} Cases counts selected persistent misrankings; Confirmed counts cases repaired on validation and held-out prompts at any search level. R--C and C--C count reference--candidate and candidate--candidate cases. Res., Mean, and Zero count cases by their original resampling, mean, or zero intervention. Each of these two partitions sums to Cases. Numeric tasks are InterpBench IDs. The cohort contains 100 distinct pairs across fourteen tasks.}\label{tab:s5allocation}
\footnotesize
\setlength{\tabcolsep}{3pt}
\renewcommand{\arraystretch}{1.08}
\begin{tabular}{lrrrrrrr}
\toprule
Task & Cases & Confirmed & R--C & C--C & Res. & Mean & Zero\\
\midrule
\multicolumn{8}{l}{\emph{Human suite}}\\
IOI & 13 & 12 & 8 & 5 & 10 & 3 & 0\\
Greater-Than & 12 & 12 & 5 & 7 & 3 & 2 & 7\\
Docstring & 13 & 13 & 5 & 8 & 9 & 4 & 0\\
Acronym & 12 & 12 & 5 & 7 & 4 & 3 & 5\\
\multicolumn{8}{l}{\emph{InterpBench suite}}\\
113 & 5 & 4 & 0 & 5 & 4 & 1 & 0\\
97 & 2 & 2 & 0 & 2 & 0 & 1 & 1\\
2 & 5 & 5 & 4 & 1 & 0 & 2 & 3\\
82 & 5 & 5 & 1 & 4 & 4 & 0 & 1\\
111 & 9 & 9 & 6 & 3 & 7 & 1 & 1\\
45 & 5 & 5 & 3 & 2 & 3 & 2 & 0\\
58 & 5 & 5 & 3 & 2 & 4 & 0 & 1\\
93 & 4 & 4 & 0 & 4 & 0 & 1 & 3\\
103 & 5 & 4 & 4 & 1 & 2 & 1 & 2\\
25 & 5 & 4 & 3 & 2 & 0 & 4 & 1\\
\midrule
All cases & 100 & 96 & 47 & 53 & 50 & 25 & 25\\
\bottomrule
\end{tabular}
\end{table}

\FloatBarrier

We use held-out baseline scores to ensure that each selected failure persists beyond validation. The repair rate therefore describes these 100 persistent failures. Once we begin restoration search, we use only validation scores to choose subsets; we use held-out scores to test whether those same restorations repair the ranking on separate prompts.

\subsection{Eligible groups and recipient-intact restoration}
For each pair, we keep $B$ and $W$ fixed as the higher-Q and lower-Q circuits. We allow restoration at connections excluded by either mask and group those connections by their receiving node. Let $\mathcal{G}$ denote this original eligible group set and $G=|\mathcal{G}|$. A group may contain several excluded incoming connections, so we record both the number of groups and the number of connections restored.

For each recipient prompt, we cache the signals from its intact-model execution. When we restore a selected connection, we replace its original donor, mean, or zero signal with the corresponding recipient-intact signal. We do this only where the circuit excludes the connection; retained connections continue to compute normally. We leave the masks, weights, and all other intervention signals unchanged, including the original donor assignments.

On held-out prompts, we apply the same restoration subset using each recipient's own intact cache and its original held-out donors. We continue to orient the pair by its original, unrestored Q. We do not recompute Q for the repaired executions: our question is whether restoring context makes faithfulness prefer the originally better circuit, not whether the restored hybrid is itself a better discovered circuit.

\subsection{Validation-guided search at five restoration levels}
Before searching, we confirm that KL prefers the lower-Q circuit on both splits. To count a repair, we require it to prefer $B$ beyond the same numerical tolerance used throughout the paper. For scores $F(B)$ and $F(W)$ on either split, we define
\begin{equation}
\epsilon(F(B),F(W))=10^{-6}+10^{-6}\max(|F(B)|,|F(W)|).
\end{equation}
We count a subset as a confirmed repair only when $F(B)-F(W)>\epsilon(F(B),F(W))$ on both validation and held-out prompts. Reaching a tie is not enough.

We search at fractions $f\in\{0.8,0.4,0.2,0.1,0.05\}$ in descending order, with ten predetermined attempts at each level:
\begin{enumerate}
\item We begin with the original eligible group set $\mathcal{G}$. At level $f$, we request approximately $fG$ groups, rounded to an integer and bounded below by one. We always measure the percentage relative to the original $G$, including when we sample from a successful parent subset.
\item We evaluate all ten subsets at that level on validation. We use the first validation-successful subset as the parent for the next level; later successes do not change that choice.
\item If none succeeds, we sample the next level from $\mathcal{G}$ again. If this fallback finds a repair, we resume refinement inside its first successful subset.
\item We test every validation-successful subset unchanged on held-out prompts, including subsets not selected as parents. We do not use the held-out results to choose parents, alter random draws, or stop the search.
\item We continue through 5\%, allowing at most 50 trials per case. We count the case as repaired if any validation-successful subset also repairs the held-out ranking, regardless of what happens at later levels.
\end{enumerate}

This search lets us find a broad restoration and then ask whether a smaller one can also work. It also allows a smaller subset to succeed when no larger subset does. Because groups contain different numbers of excluded connections, the nominal fraction does not specify the exact number of restored sites. We report both counts in Table~\ref{tab:s5searchdetails}, together with validation and held-out outcomes. We seek a sufficient repair within the budget; we do not test minimality or location specificity, and we do not run matched random controls.
\begin{table*}[t]
\centering
\caption{\textbf{Restoration outcomes and subset sizes by level.} Level percentages refer to the original eligible receiving-node groups, with ten restoration attempts per level. Validation counts cases with a validation-successful subset; Test confirmed counts cases where the same subset also repairs the held-out ranking, each out of 100. Groups restored and Sites restored give median [25th, 75th percentiles] of receiving-node groups and excluded connections in the first confirmed subset per case. The first validation success determines the next parent, independently of test results.}\label{tab:s5searchdetails}
\scriptsize
\setlength{\tabcolsep}{3pt}
\begin{tabular}{lrrrr}
\toprule
Level & Validation & Test confirmed & Groups restored & Sites restored\\
\midrule
80\% & 92/100 & 92/100 & 10.0 [9.0, 24.8] & 283.0 [137.8, 6597.0]\\
40\% & 88/100 & 87/100 & 5.0 [4.0, 10.0] & 128.0 [69.5, 601.0]\\
20\% & 84/100 & 82/100 & 2.0 [2.0, 5.0] & 83.5 [24.2, 398.5]\\
10\% & 76/100 & 66/100 & 1.0 [1.0, 2.0] & 31.0 [12.0, 186.2]\\
5\% & 72/100 & 60/100 & 1.0 [1.0, 1.0] & 17.0 [8.0, 64.8]\\
\bottomrule
\end{tabular}
\end{table*}

\subsection{Confirmed repairs and fixed behavioral deficits}
We find at least one held-out-confirmed repair in 96 of the 100 persistent failures: 49/50 Human cases and 47/50 InterpBench cases. At the five restoration levels, we confirm 92, 87, 82, 66, and 60 cases, respectively. These counts overlap because one case can be repaired at several levels. The first four levels already repair 95 cases; the 5\% level adds one more. If we test only the first validation-successful trial at each level, we confirm 95/100 across all five levels. Testing the other validation-successful trials finds the remaining repair without using held-out scores to direct the search.

For each case, we define $\Delta Q_{0}=Q_{0}(B)-Q_{0}(W)>0$ using the original, unrestored executions. We use this gap to select and orient the pair; its mean is 3.36 pp across the cohort. Restoration can change both circuits' outputs, but the outcome we measure is whether the score now prefers the originally better circuit. We count each repaired case once in the 96/100 total, even if several subsets work, and retain all four unconfirmed cases in the relevant denominators.

We give the exact counts behind Table~\ref{tab:s5summary} in Table~\ref{tab:s5counts}, every case's level outcomes in Table~\ref{tab:s5catalog}, and the breakdown by suite, intervention, pair type, and method in Table~\ref{tab:s5breakdown}. Together, these tables show where restoration succeeds within the selected cohort and where our search does not find a confirmed repair.
\begin{table}[t]
\centering
\caption{\textbf{Context restoration repairs persistent misrankings.} Rows group cases by suite, original intervention, and pair type; R--C is reference--candidate and C--C is candidate--candidate. $n$ is the number of selected cases in that row. Level percentages refer to the original eligible receiving-node groups, with ten restoration attempts per level. A confirmed repair corrects the original KL misranking on both validation and held-out prompts. Entries are confirmed cases divided by $n$, expressed as percentages. Any level counts the union of repaired cases; level-specific counts overlap.}\label{tab:s5summary}
\begingroup
\small
\setlength{\tabcolsep}{3pt}
\renewcommand{\arraystretch}{1.12}
\resizebox{\textwidth}{!}{%
\begin{tabular*}{\linewidth}{@{\extracolsep{\fill}}llrrrrrrr@{}}
\toprule
 & & & \multicolumn{6}{c}{Confirmed repair rates (\%)}\\
\cmidrule(l){4-9}
Intervention & Pair & $n$ & 80\% & 40\% & 20\% & 10\% & 5\% & Any level\\
\midrule
\multicolumn{9}{@{}l}{\emph{Human suite (50 cases)}}\\
\addlinespace[2pt]
\multirow{2}{*}{Resampling} & R--C & 12 & 83.3 & 66.7 & 58.3 & 41.7 & 41.7 & 91.7\\
 & C--C & 14 & 100.0 & 100.0 & 100.0 & 78.6 & 64.3 & 100.0\\
\addlinespace[2pt]
\multirow{2}{*}{Mean} & R--C & 6 & 100.0 & 83.3 & 50.0 & 50.0 & 50.0 & 100.0\\
 & C--C & 6 & 100.0 & 83.3 & 83.3 & 83.3 & 50.0 & 100.0\\
\addlinespace[2pt]
\multirow{2}{*}{Zero} & R--C & 5 & 100.0 & 100.0 & 100.0 & 100.0 & 80.0 & 100.0\\
 & C--C & 7 & 100.0 & 100.0 & 85.7 & 57.1 & 71.4 & 100.0\\
\midrule
\multicolumn{9}{@{}l}{\emph{InterpBench suite (50 cases)}}\\
\addlinespace[2pt]
\multirow{2}{*}{Resampling} & R--C & 11 & 100.0 & 100.0 & 100.0 & 72.7 & 72.7 & 100.0\\
 & C--C & 13 & 84.6 & 84.6 & 92.3 & 76.9 & 76.9 & 100.0\\
\addlinespace[2pt]
\multirow{2}{*}{Mean} & R--C & 6 & 83.3 & 100.0 & 100.0 & 83.3 & 66.7 & 100.0\\
 & C--C & 7 & 71.4 & 71.4 & 57.1 & 57.1 & 57.1 & 71.4\\
\addlinespace[2pt]
\multirow{2}{*}{Zero} & R--C & 7 & 100.0 & 85.7 & 71.4 & 42.9 & 28.6 & 100.0\\
 & C--C & 6 & 83.3 & 66.7 & 66.7 & 50.0 & 50.0 & 83.3\\
\midrule
\multicolumn{2}{@{}l}{\emph{All 100 cases}} & 100 & 92.0 & 87.0 & 82.0 & 66.0 & 60.0 & 96.0\\
\bottomrule
\end{tabular*}%
}
\endgroup
\end{table}

\begin{table*}[t]
\centering
\caption{\textbf{Exact confirmed-repair counts.} Rows group cases by suite, original intervention, and pair type (R--C: reference--candidate; C--C: candidate--candidate). Resampling uses the original donor assignments. Level percentages refer to the original eligible receiving-node groups, with ten restoration attempts per level. A confirmed repair corrects the original KL misranking on both validation and held-out prompts. Entries are confirmed repairs / selected cases in that row. Any level counts each repaired case once across levels; level-specific counts overlap.}\label{tab:s5counts}
\scriptsize
\setlength{\tabcolsep}{3pt}
\begin{tabular}{llrrrrrr}
\toprule
Intervention & Pair & 80\% & 40\% & 20\% & 10\% & 5\% & Any level\\
\midrule
\multicolumn{8}{l}{\emph{Human suite}}\\
Resampling & R--C & 10/12 & 8/12 & 7/12 & 5/12 & 5/12 & 11/12\\
Resampling & C--C & 14/14 & 14/14 & 14/14 & 11/14 & 9/14 & 14/14\\
Mean & R--C & 6/6 & 5/6 & 3/6 & 3/6 & 3/6 & 6/6\\
Mean & C--C & 6/6 & 5/6 & 5/6 & 5/6 & 3/6 & 6/6\\
Zero & R--C & 5/5 & 5/5 & 5/5 & 5/5 & 4/5 & 5/5\\
Zero & C--C & 7/7 & 7/7 & 6/7 & 4/7 & 5/7 & 7/7\\
\multicolumn{8}{l}{\emph{InterpBench suite}}\\
Resampling & R--C & 11/11 & 11/11 & 11/11 & 8/11 & 8/11 & 11/11\\
Resampling & C--C & 11/13 & 11/13 & 12/13 & 10/13 & 10/13 & 13/13\\
Mean & R--C & 5/6 & 6/6 & 6/6 & 5/6 & 4/6 & 6/6\\
Mean & C--C & 5/7 & 5/7 & 4/7 & 4/7 & 4/7 & 5/7\\
Zero & R--C & 7/7 & 6/7 & 5/7 & 3/7 & 2/7 & 7/7\\
Zero & C--C & 5/6 & 4/6 & 4/6 & 3/6 & 3/6 & 5/6\\
\midrule
All &  & 92/100 & 87/100 & 82/100 & 66/100 & 60/100 & 96/100\\
\bottomrule
\end{tabular}
\end{table*}

\FloatBarrier
\FloatBarrier
\begingroup\scriptsize\setlength{\tabcolsep}{2pt}
\begin{longtable}{llllrrrrrr}
\caption{\textbf{All selected restoration cases.} Each row is a distinct circuit pair under its original intervention. Task names the behavior or InterpBench ID; Method names the discovery algorithm; Pair is reference--candidate (R--C) or candidate--candidate (C--C). $\Delta Q_0$ is the original held-out behavioral deficit, before restoration, in percentage points (pp). The 80\%--5\% columns give nominal fractions of eligible receiving-node groups restored: Y means at least one subset repairs both validation and held-out rankings; N means no confirmed repair within ten attempts. A case can have several Y entries.}\label{tab:s5catalog}\\
\toprule
Task & Method & Intervention & Pair & $\Delta Q_{0}$ (pp) & 80\% & 40\% & 20\% & 10\% & 5\%\\
\midrule\endfirsthead
\caption[]{Selected restoration cases (continued).}\\
\toprule
Task & Method & Intervention & Pair & $\Delta Q_{0}$ (pp) & 80\% & 40\% & 20\% & 10\% & 5\%\\
\midrule\endhead
\bottomrule\endfoot
\bottomrule\endlastfoot
\multicolumn{10}{l}{\emph{Human suite}}\\
IOI & EAP & Mean & C--C & 1.67 & Y & Y & Y & Y & N\\
IOI & EAP-IG & Resampling & R--C & 1.33 & N & N & N & N & N\\
IOI & EAP-IG & Resampling & R--C & 3.00 & Y & N & N & N & N\\
IOI & EAP-IG & Resampling & R--C & 0.67 & Y & N & N & N & N\\
IOI & EAP-IG & Resampling & R--C & 1.33 & N & Y & Y & N & N\\
IOI & EAP-IG & Resampling & R--C & 3.00 & Y & Y & Y & N & N\\
IOI & EAP-IG & Resampling & R--C & 1.33 & Y & N & N & N & N\\
IOI & EAP-IG & Resampling & C--C & 1.00 & Y & Y & Y & Y & N\\
IOI & EAP-IG & Mean & R--C & 0.67 & Y & N & N & N & N\\
IOI & ACDC & Resampling & C--C & 0.67 & Y & Y & Y & Y & Y\\
IOI & ACDC & Mean & R--C & 5.33 & Y & Y & N & Y & Y\\
IOI & Edge-SP & Resampling & C--C & 4.33 & Y & Y & Y & Y & Y\\
IOI & Edge-SP & Resampling & C--C & 1.67 & Y & Y & Y & Y & Y\\
Greater-Than & EAP & Resampling & C--C & 0.67 & Y & Y & Y & N & N\\
Greater-Than & EAP & Resampling & C--C & 1.67 & Y & Y & Y & Y & Y\\
Greater-Than & EAP & Zero & R--C & 3.67 & Y & Y & Y & Y & N\\
Greater-Than & EAP & Zero & R--C & 2.67 & Y & Y & Y & Y & Y\\
Greater-Than & EAP & Zero & C--C & 2.67 & Y & Y & Y & N & Y\\
Greater-Than & EAP & Zero & C--C & 0.67 & Y & Y & Y & Y & Y\\
Greater-Than & EAP-IG & Resampling & C--C & 0.67 & Y & Y & Y & Y & N\\
Greater-Than & EAP-IG & Mean & C--C & 0.67 & Y & Y & Y & Y & Y\\
Greater-Than & EAP-IG & Mean & C--C & 1.33 & Y & Y & Y & Y & N\\
Greater-Than & EAP-IG & Zero & R--C & 1.00 & Y & Y & Y & Y & Y\\
Greater-Than & ACDC & Zero & R--C & 1.00 & Y & Y & Y & Y & Y\\
Greater-Than & Edge-SP & Zero & R--C & 0.67 & Y & Y & Y & Y & Y\\
Docstring & EAP & Mean & C--C & 1.00 & Y & Y & Y & Y & Y\\
Docstring & EAP-IG & Resampling & R--C & 3.00 & Y & Y & Y & Y & Y\\
Docstring & EAP-IG & Resampling & R--C & 1.33 & Y & Y & N & N & N\\
Docstring & EAP-IG & Resampling & C--C & 0.67 & Y & Y & Y & N & N\\
Docstring & EAP-IG & Resampling & C--C & 3.67 & Y & Y & Y & Y & Y\\
Docstring & ACDC & Resampling & C--C & 0.67 & Y & Y & Y & Y & Y\\
Docstring & ACDC & Resampling & C--C & 0.67 & Y & Y & Y & N & N\\
Docstring & ACDC & Resampling & C--C & 1.67 & Y & Y & Y & Y & Y\\
Docstring & Edge-SP & Resampling & C--C & 3.00 & Y & Y & Y & Y & Y\\
Docstring & Edge-SP & Resampling & C--C & 6.00 & Y & Y & Y & Y & Y\\
Docstring & Edge-SP & Mean & R--C & 4.67 & Y & Y & Y & N & Y\\
Docstring & Edge-SP & Mean & R--C & 9.67 & Y & Y & Y & N & N\\
Docstring & Edge-SP & Mean & R--C & 10.67 & Y & Y & N & Y & Y\\
Acronym & EAP & Resampling & R--C & 2.33 & Y & Y & Y & Y & Y\\
Acronym & EAP-IG & Zero & C--C & 2.00 & Y & Y & Y & N & N\\
Acronym & ACDC & Resampling & R--C & 2.78 & Y & Y & Y & Y & Y\\
Acronym & ACDC & Resampling & R--C & 1.78 & Y & Y & Y & Y & Y\\
Acronym & ACDC & Resampling & R--C & 2.44 & Y & Y & Y & Y & Y\\
Acronym & Edge-SP & Mean & R--C & 49.56 & Y & Y & Y & Y & N\\
Acronym & Edge-SP & Mean & C--C & 0.89 & Y & N & N & N & N\\
Acronym & Edge-SP & Mean & C--C & 20.33 & Y & Y & Y & Y & Y\\
Acronym & Edge-SP & Zero & C--C & 20.67 & Y & Y & Y & Y & Y\\
Acronym & Edge-SP & Zero & C--C & 6.33 & Y & Y & Y & Y & Y\\
Acronym & Edge-SP & Zero & C--C & 2.22 & Y & Y & Y & Y & Y\\
Acronym & Edge-SP & Zero & C--C & 66.33 & Y & Y & N & N & N\\
\multicolumn{10}{l}{\emph{InterpBench suite}}\\
113 & EAP & Resampling & C--C & 0.93 & Y & Y & Y & Y & Y\\
113 & EAP & Resampling & C--C & 0.81 & Y & Y & Y & Y & Y\\
113 & EAP & Resampling & C--C & 1.48 & Y & Y & Y & Y & Y\\
113 & EAP & Resampling & C--C & 1.67 & Y & Y & Y & Y & Y\\
113 & ACDC & Mean & C--C & 1.11 & N & N & N & N & N\\
97 & EAP & Zero & C--C & 0.59 & Y & Y & Y & Y & Y\\
97 & ACDC & Mean & C--C & 2.96 & Y & Y & N & N & N\\
2 & EAP & Zero & C--C & 8.93 & Y & Y & Y & Y & Y\\
2 & EAP-IG & Zero & R--C & 2.22 & Y & Y & Y & Y & N\\
2 & ACDC & Mean & R--C & 1.44 & Y & Y & Y & Y & N\\
2 & ACDC & Mean & R--C & 0.96 & Y & Y & Y & Y & Y\\
2 & Edge-SP & Zero & R--C & 1.74 & Y & Y & N & N & N\\
82 & EAP & Resampling & C--C & 1.44 & Y & Y & Y & Y & Y\\
82 & EAP-IG & Zero & R--C & 0.56 & Y & Y & Y & Y & Y\\
82 & ACDC & Resampling & C--C & 0.56 & N & N & N & N & Y\\
82 & ACDC & Resampling & C--C & 0.56 & N & N & Y & Y & Y\\
82 & Edge-SP & Resampling & C--C & 0.59 & Y & Y & Y & Y & N\\
111 & EAP & Resampling & R--C & 0.93 & Y & Y & Y & Y & Y\\
111 & EAP & Resampling & R--C & 0.52 & Y & Y & Y & Y & Y\\
111 & EAP-IG & Mean & C--C & 0.96 & Y & Y & Y & Y & Y\\
111 & ACDC & Resampling & R--C & 0.81 & Y & Y & Y & Y & Y\\
111 & ACDC & Resampling & R--C & 0.93 & Y & Y & Y & Y & Y\\
111 & ACDC & Resampling & R--C & 1.22 & Y & Y & Y & Y & Y\\
111 & ACDC & Resampling & C--C & 0.52 & Y & Y & Y & N & N\\
111 & ACDC & Resampling & C--C & 0.52 & Y & Y & Y & N & N\\
111 & Edge-SP & Zero & R--C & 2.19 & Y & Y & Y & Y & Y\\
45 & EAP & Resampling & C--C & 1.11 & Y & Y & Y & Y & Y\\
45 & EAP-IG & Mean & R--C & 1.37 & N & Y & Y & Y & Y\\
45 & ACDC & Resampling & R--C & 0.56 & Y & Y & Y & Y & Y\\
45 & ACDC & Resampling & R--C & 0.63 & Y & Y & Y & N & N\\
45 & Edge-SP & Mean & C--C & 0.52 & Y & Y & Y & Y & Y\\
58 & EAP & Resampling & R--C & 0.93 & Y & Y & Y & N & N\\
58 & EAP-IG & Zero & R--C & 1.22 & Y & Y & Y & N & N\\
58 & ACDC & Resampling & C--C & 1.22 & Y & Y & Y & Y & Y\\
58 & ACDC & Resampling & C--C & 1.44 & Y & Y & Y & Y & Y\\
58 & Edge-SP & Resampling & R--C & 0.74 & Y & Y & Y & N & N\\
93 & EAP & Zero & C--C & 6.85 & Y & N & N & N & N\\
93 & EAP & Zero & C--C & 6.85 & Y & Y & Y & N & N\\
93 & ACDC & Mean & C--C & 0.93 & Y & Y & Y & Y & Y\\
93 & Edge-SP & Zero & C--C & 0.81 & Y & Y & Y & Y & Y\\
103 & EAP & Mean & C--C & 0.63 & N & N & N & N & N\\
103 & EAP-IG & Zero & R--C & 0.59 & Y & N & N & N & N\\
103 & ACDC & Resampling & R--C & 0.67 & Y & Y & Y & Y & Y\\
103 & ACDC & Resampling & R--C & 0.74 & Y & Y & Y & Y & Y\\
103 & Edge-SP & Zero & R--C & 0.52 & Y & Y & Y & N & N\\
25 & EAP & Mean & R--C & 0.63 & Y & Y & Y & Y & Y\\
25 & EAP-IG & Mean & C--C & 0.52 & Y & Y & Y & Y & Y\\
25 & ACDC & Zero & C--C & 0.89 & N & N & N & N & N\\
25 & Edge-SP & Mean & R--C & 0.63 & Y & Y & Y & Y & Y\\
25 & Edge-SP & Mean & R--C & 0.56 & Y & Y & Y & N & N\\
\end{longtable}
\endgroup

\FloatBarrier
\begingroup\scriptsize\setlength{\tabcolsep}{2pt}
\begin{longtable}{llrrrrrr}
\caption{\textbf{Restoration outcomes by cohort subgroup.} Grouping identifies suite, original intervention, pair type, or discovery method; Subgroup names its value. R--C is reference--candidate and C--C is candidate--candidate. Level percentages refer to the original eligible receiving-node groups, with ten restoration attempts per level. A confirmed repair corrects the original KL misranking on both validation and held-out prompts. Entries are confirmed repairs / selected cases in the subgroup. Any level is their union; levels overlap and the same cases appear in different grouping categories.}\label{tab:s5breakdown}\\
\toprule
Grouping & Subgroup & 80\% & 40\% & 20\% & 10\% & 5\% & Any level\\
\midrule\endfirsthead
\caption[]{Restoration outcomes by subgroup (continued).}\\
\toprule
Grouping & Subgroup & 80\% & 40\% & 20\% & 10\% & 5\% & Any level\\
\midrule\endhead
\bottomrule\endfoot
\bottomrule\endlastfoot
Suite & Human & 48/50 & 44/50 & 40/50 & 33/50 & 29/50 & 49/50\\
Suite & InterpBench & 44/50 & 43/50 & 42/50 & 33/50 & 31/50 & 47/50\\
Intervention & Resampling & 46/50 & 44/50 & 44/50 & 34/50 & 32/50 & 49/50\\
Intervention & Mean & 22/25 & 21/25 & 18/25 & 17/25 & 14/25 & 23/25\\
Intervention & Zero & 24/25 & 22/25 & 20/25 & 15/25 & 14/25 & 24/25\\
Pair type & R--C & 44/47 & 41/47 & 37/47 & 29/47 & 26/47 & 46/47\\
Pair type & C--C & 48/53 & 46/53 & 45/53 & 37/53 & 34/53 & 50/53\\
Method & EAP & 23/24 & 22/24 & 22/24 & 18/24 & 17/24 & 23/24\\
Method & EAP-IG & 21/24 & 18/24 & 17/24 & 12/24 & 8/24 & 23/24\\
Method & ACDC & 24/28 & 24/28 & 23/28 & 20/28 & 20/28 & 26/28\\
Method & Edge-SP & 24/24 & 23/24 & 20/24 & 16/24 & 15/24 & 24/24\\
\end{longtable}
\endgroup

\FloatBarrier

\section{Same-pair metric overlap and agreement deficits}
\label{app:overlap}
We next ask whether different metrics fail on the same pairs. For each task, method, intervention, and pair type, we match KL, LD, and PD results by circuit identity and require the same Q values. This keeps the behavioral comparison fixed while changing the score. We classify each common pair into one of eight outcomes: all three metrics misrank; only KL+LD, KL+PD, or LD+PD misrank; only KL, LD, or PD misranks; or none does. We retain ties in the denominator. The ``None'' category includes both correct strict rankings and ties, which we separate in the accompanying analysis.

We find both shared and metric-specific failures (Table~\ref{tab:overlap}). In the Human discovery C--C pool, all three metrics misrank 1,354 comparisons, while 1,709 misrank only under KL. We also find that an absence of reversals can reflect an inability to distinguish candidates: of the 30,134 InterpBench discovery C--C comparisons classified as ``None,'' 17,130 have tied Q. We report these ties and the behavioral deficits of shared failures in Table~\ref{tab:overlapties}.

Agreement among the metrics does not guarantee a sound choice. Of the 5,809 Human controlled C--C comparisons misranked by all three, 2,378 have behavioral deficits of at least 5 pp. We retain repeated pairs across interventions in these summaries, so the pooled comparisons are dependent. In Table~\ref{tab:examples}, we show one shared failure per Human task, chosen by a fixed lexicographic rule rather than by the size of its deficit. All three metrics prefer the worse candidate in each example. These examples have modest Q gaps; the complete distributions show the larger losses as well.
\begin{table}[!tbp]
\centering
\caption{\textbf{Shared and metric-specific misrankings.} Panel / pair identifies the experiment (S3: controlled edits; S4: discovery outputs), suite, and pair type (R--C: reference--candidate; C--C: candidate--candidate). Pool counts common pair--intervention comparisons, including ties. All 3 counts failures shared by all metrics; two-metric and single-metric columns count failures of exactly those metrics. None counts comparisons with no strict misranking, including ties. These eight outcomes partition Pool.}\label{tab:overlap}
\scriptsize
\setlength{\tabcolsep}{3pt}
\begin{tabular}{lrrrrrrrrr}
\toprule
Panel / pair & Pool & All 3 & KL+LD & KL+PD & LD+PD & KL & LD & PD & None\\
\midrule
S3 \emph{Human suite} C--C & 151605 & 5809 & 3297 & 2548 & 14712 & 15283 & 4113 & 21155 & 84688\\
S3 \emph{Human suite} R--C & 1890 & 50 & 28 & 34 & 172 & 181 & 35 & 230 & 1160\\
S3 \emph{InterpBench suite} C--C & 467958 & 14864 & 5115 & 13344 & 3336 & 32687 & 15740 & 2579 & 380293\\
S3 \emph{InterpBench suite} R--C & 5307 & 155 & 52 & 161 & 42 & 305 & 196 & 33 & 4363\\
S4 \emph{Human suite} C--C & 15948 & 1354 & 963 & 592 & 1140 & 1709 & 877 & 1588 & 7725\\
S4 \emph{Human suite} R--C & 1176 & 87 & 37 & 25 & 189 & 230 & 106 & 78 & 424\\
S4 \emph{InterpBench suite} C--C & 45267 & 3019 & 570 & 2752 & 567 & 1144 & 6665 & 416 & 30134\\
S4 \emph{InterpBench suite} R--C & 3261 & 421 & 13 & 157 & 158 & 44 & 495 & 22 & 1951\\
\bottomrule
\end{tabular}
\end{table}

\begin{table}[!tbp]
\centering
\caption{\textbf{Ties and behavioral deficits in the common metric pools.} S3 and S4 denote controlled and discovery experiments; R--C and C--C denote reference--candidate and candidate--candidate pairs. Within None (no metric misranks), Q ties have indistinguishable held-out behavior; Score ties have distinct behavior but at least one tied score; Correct means all three metrics strictly prefer the better circuit. These categories partition None. Under All three misrank, Failures counts shared KL/LD/PD reversals, Median [Q1, Q3] gives the behavioral deficit\textquotesingle s median [25th, 75th percentiles] in percentage points (pp), and Deficit $\geq5$ pp counts shared failures at or above that threshold.}\label{tab:overlapties}
\resizebox{\textwidth}{!}{%
\begin{tabular}{lrrrrrrr}
\toprule
 & \multicolumn{4}{c}{No metric misranks} & \multicolumn{3}{c}{All three misrank}\\
\cmidrule(lr){2-5}\cmidrule(lr){6-8}
Panel / pair & None & Q ties & Score ties & Correct & Failures & Median [Q1, Q3] (pp) & Deficit $\geq5$ pp\\
\midrule
S3 \emph{Human suite} C--C & 84688 & 7812 & 442 & 76434 & 5809 & 3.00 [0.67, 15.33] & 2378\\
S3 \emph{Human suite} R--C & 1160 & 177 & 6 & 977 & 50 & 1.33 [0.42, 4.33] & 12\\
S3 \emph{InterpBench suite} C--C & 380293 & 56988 & 4523 & 318782 & 14864 & 4.44 [0.78, 7.11] & 6851\\
S3 \emph{InterpBench suite} R--C & 4363 & 732 & 125 & 3506 & 155 & 3.52 [0.30, 7.15] & 74\\
S4 \emph{Human suite} C--C & 7725 & 2466 & 39 & 5220 & 1354 & 1.00 [0.33, 2.33] & 144\\
S4 \emph{Human suite} R--C & 424 & 54 & 0 & 370 & 87 & 0.67 [0.33, 4.50] & 18\\
S4 \emph{InterpBench suite} C--C & 30134 & 17130 & 1137 & 11867 & 3019 & 0.22 [0.07, 0.70] & 150\\
S4 \emph{InterpBench suite} R--C & 1951 & 192 & 279 & 1480 & 421 & 1.04 [0.48, 1.56] & 3\\
\bottomrule
\end{tabular}
}%
\end{table}

\begin{table}[!tbp]
\centering
\caption{\textbf{Examples misranked by all three metrics.} For each Human task, we choose the lexicographically first controlled pair shared by KL, LD, and PD. Circuit 1 and Circuit 2 are identifiers, not quality rankings: \texttt{k} is the number of swapped components and \texttt{s} is the generation seed. $\Delta Q$ is the better circuit\textquotesingle s held-out advantage in percentage points (pp). Each score gap is worse minus better, so positive gaps favor the worse circuit; KL gaps use natural-log units, LD gaps logit units, and PD gaps probability units.}\label{tab:examples}
\scriptsize
\setlength{\tabcolsep}{3pt}
\begin{tabular}{llllrrrr}
\toprule
Task & Intervention & Circuit 1 & Circuit 2 & $\Delta Q$ (pp) & KL gap & LD gap & PD gap\\
\midrule
IOI & Mean & k193\_s0 & k193\_s1 & 0.33 & 0.206 & 0.148 & 0.000852\\
Greater-Than & Mean & k118\_s16 & k118\_s17 & 0.67 & 0.00226 & 0.0683 & 0.0176\\
Docstring & Mean & k12\_s0 & k12\_s3 & 1.00 & 0.00648 & 0.00267 & 0.000209\\
Acronym & Mean & k1\_s12 & k1\_s17 & 0.33 & 0.0427 & 0.0802 & 0.00557\\
\bottomrule
\end{tabular}
\end{table}

\FloatBarrier

\end{document}